\documentclass[acmtog]{acmart}

\acmSubmissionID{papers\_642}

\AtBeginDocument{%
  \providecommand\BibTeX{{%
    \normalfont B\kern-0.5em{\scshape i\kern-0.25em b}\kern-0.8em\TeX}}}

\setcopyright{rightsretained}
\acmJournal{TOG}
\acmYear{2023} \acmVolume{42} \acmNumber{4} \acmArticle{1} \acmMonth{8} \acmPrice{15.00} \acmDOI{10.1145/3592429}

\usepackage{xcolor}
\usepackage{soul}
\usepackage{caption}
\usepackage{subcaption}
\usepackage{xspace}
\usepackage{cancel}
\usepackage{mathtools}
\usepackage{booktabs}
\usepackage{graphicx}
\graphicspath{{./fig/}{./fig/background}{./fig/goal0}{./fig/goal1}{./fig/goal2}{./fig/interference}{./fig/teaser}{./fig/real}{./fig/overview}}
\usepackage{pifont}
\usepackage{amsmath}
\usepackage{tabularx}
\usepackage[inline]{enumitem}
\usepackage{stmaryrd}
\usepackage[squaren]{SIunits}

\AtBeginDocument{\colorlet{defaultcolor}{.}}

\begin{document}
\newcommand{\fref}[1]{Figure~\ref{#1}}
\newcommand{\tref}[1]{Table~\ref{#1}}
\newcommand{\eref}[1]{Equation~\ref{#1}}
\newcommand{\eeref}[2]{Equations~\ref{#1} and \ref{#2}}
\newcommand{\cref}[1]{Chapter~\ref{#1}}
\newcommand{\sref}[1]{Section~\ref{#1}}
\newcommand{\ssref}[2]{Sections~\ref{#1}~and~\ref{#2}}
\newcommand{\aref}[1]{Annex~\ref{#1}}

\newcommand{\crhide}[1]{#1}
\newcommand{\phide}[1]{#1}

\newcommand{\draft}[1]{#1}

\definecolor{red}{rgb}{0.8,0,0}
\definecolor{purered}{rgb}{1,0,0}
\definecolor{pink}{rgb}{0.9,0,0.9}
\definecolor{darkred}{rgb}{0.6,0,0}
\definecolor{green}{rgb}{0.0,0.5,0}
\definecolor{blue}{rgb}{0,0,0.75}
\definecolor{darkblue}{rgb}{0,0,0.55}
\definecolor{lightcyan}{rgb}{0.5,0.7,0.7}
\definecolor{orange}{rgb}{0.9,0.3,0.1}
\definecolor{purple}{rgb}{0.6,0.0,0.6}
\definecolor{cyan}{rgb}{0.0,0.7,0.7}
\definecolor{darkgray}{rgb}{0.4,0.4,0.4}
\definecolor{bronze}{rgb}{0.7, 0.4, 0.18}
\definecolor{dorange}{rgb}{0.75, 0.4, 0.0}
\definecolor{black}{rgb}{0.0,0.0,0.0}

\newcommand{\new}[1]{\textcolor{defaultcolor}{#1}}
\newcommand{\neww}[1]{\textcolor{defaultcolor}{#1}}
\newcommand{\newww}[1]{\textcolor{defaultcolor}{#1}}
\newcommand{\NEW}[1]{\textcolor{defaultcolor}{#1}}
\newcommand{\NEWW}[1]{\textcolor{defaultcolor}{#1}}


\newcommand{\exptAvailable}[1]{\colortxt{green}{#1}}
\newcommand{\exptTentative}[1]{\colortxt{red}{#1}}
\newcommand{\rarr}{\rightarrow}
\newcommand{\larr}{\leftarrow}
\newcommand{\transpose}{{\scriptstyle\top}}
\renewcommand{\transpose}{{\intercal}}
\newcommand{\minv}[1]{{#1}^{-1}}
\newcommand{\Fspace}[1]{\widehat{#1}}
\newcommand{\Fourier}[1]{\mathcal{F}\left\{#1\right\}}
\newcommand{\invFourier}[1]{\mathcal{F}^{-1}\left\{#1\right\}}
\newcommand{\mat}[1]{{\mathbf{#1}}}
\newcommand{\reparam}[1]{\mathrm{#1}}
\newcommand{\freq}{{\Omega}}
\newcommand{\fq}{{\freq}}
\newcommand{\wl}{{\lambda}}
\newcommand{\conv}{\ast}
\newcommand\given[1][]{\:#1\vert\:}

\newcommand{\cmark}{\ding{51}}%
\newcommand{\xmark}{\ding{55}}%
\newcommand{\tododone}{\cmark\xspace}%
\newcommand{\todofail}{\xmark\xspace}%
\newcommand{\enumstrength}{S.\arabic*}%
\newcommand{\enumproblem}{P.\arabic*}%

\newcommand*{\pmapping}[0]{\emph{photon mapping}}
\newcommand*{\Pmapping}[0]{\emph{Photon mapping}}
\newcommand*{\PMapping}[0]{\emph{Photon Mapping}}
\newcommand*{\pbeams}[0]{\emph{photon beams}}
\newcommand*{\Pbeams}[0]{\emph{Photon beams}}
\newcommand*{\PBeams}[0]{\emph{Photon Beams}}
\newcommand*{\Ptracing}[0]{\emph{Path tracing}}
\newcommand*{\VPTracing}[0]{\emph{Volumetric Path Tracing}}
\newcommand*{\PTracing}[0]{\emph{Path Tracing}}
\newcommand*{\ptracing}[0]{\emph{path tracing}}
\newcommand*{\Bptracing}[0]{\emph{Bidirectional path tracing}}
\newcommand*{\bptracing}[0]{\emph{bidirectional path tracing}}
\newcommand*{\Metropolis}[0]{\emph{Metropolis light transport}}
\newcommand*{\brdf}[0]{\emph{BRDF}}

\newcommand{\Tr}{{T_r}}
\newcommand{\alb}{{_\Lambda}}
\newcommand{\scalb}{\alpha}
\newcommand{\crossscat}{\kappa_s}
\newcommand{\crossabs}{\kappa_a}
\newcommand*{\pf}[0]{\rho}
\newcommand*{\coefext}[0]{\sigma_t}
\newcommand*{\coefextx}[0]{\ext(x)}
\newcommand*{\coefabs}[0]{\sigma_a}
\newcommand*{\coefabsx}[0]{\abs(x)}
\newcommand*{\coefscat}[0]{\sigma_s}
\newcommand*{\coefscatx}[0]{\sca(x)}
\newcommand*{\mcoefext}[0]{$\coefext$}
\newcommand*{\mcoefextx}[0]{$\coefextx$}
\newcommand*{\mcoefabs}[0]{$\coefabs$}
\newcommand*{\mcoefabsx}[0]{$\coefabsx$}
\newcommand*{\mcoefscat}[0]{$\coefscat$}
\newcommand*{\mcoefscatx}[0]{$\coefscatx$}
\newcommand*{\phasefunction}[0]{p(x, \vec{\omega}^\prime, \vec{\omega})}
\newcommand*{\ior}[0]{\eta}
\newcommand{\diff}{\mathrm{d}}
\newcommand{\bvec}[1]{\mathbf{#1}}
\newcommand{\omegav}{\vec{\omega}}
\newcommand{\omegaout}{\omegav_o}
\newcommand{\omegain}{\omegav_i}
\newcommand{\momegain}{$\omegain$}
\newcommand{\ppwr}{\Phi}
\newcommand{\eqbreak}{\nonumber \\}
\newcommand{\x}{\ensuremath{\mathbf{x}}\xspace}
\newcommand{\y}{\ensuremath{\mathbf{y}}\xspace}
\newcommand{\z}{\ensuremath{\mathbf{z}}\xspace}
\newcommand{\rvec}{\vec{\ensuremath{\mathbf{r}}}\xspace}
\newcommand{\rlen}{\ensuremath{\mathrm{r}}\xspace}
\newcommand{\xpr}{\ensuremath{\mathbf{x}^\prime}\xspace}
\newcommand{\ym}{{\y}}
\newcommand{\zm}{{\z}}
\newcommand{\VRegion}{\ensuremath{\aleph}\xspace}
\newcommand{\xs}{\ensuremath{\x_{s}}\xspace} 
\newcommand{\ys}{\ensuremath{\y_{\! s}}\xspace}
\newcommand{\yw}{\ensuremath{\hat{\y}}\xspace}
\newcommand{\mpar}{r}
\newcommand{\pkern}{\Omega}
\newcommand{\optpath}{\Pi}

\newcommand{\stransportImage}{\mathbf{i}}
\newcommand{\stransportMatrix}{\mathbf{T}}
\newcommand{\stransportMatrixTransient}{\mathbf{H}}
\newcommand{\stransportSources}{\mathbf{p}}

\newcommand{\svtransportImage}{\mathbf{i_v}}
\newcommand{\svtransportMatrix}{\stransportMatrix}
\newcommand{\svtransportSources}{\mathbf{p_v}}

\let\norm\undefined
\DeclarePairedDelimiter{\norm}{\lvert}{\rvert}
\DeclarePairedDelimiter{\Norm}{\lVert}{\rVert}
\newcommand{\pftime}{t}
\newcommand{\planeC}{S}
\newcommand{\planeP}{L}
\newcommand{\xbf}{\mathbf{x}}
\newcommand{\xp}{\mathbf{x}_p} 
\newcommand{\xl}{\mathbf{x}_l} 
\newcommand{\xlG}{\xl^G} 
\newcommand{\xlM}{\xl^M} 
\newcommand{\xlGM}{\xl^{GM}} 
\newcommand{\xlMG}{\xl^{MG}} 
\newcommand{\xlp}{\xl^\prime} 
\newcommand{\xlpp}{\xl^{\prime\prime}} 
\newcommand{\vvec}{\vec{\mathbf{v}}}
\newcommand{\dxp}{\diff \xp}
\newcommand{\xc}{\mathbf{x}_c}
\newcommand{\dxc}{\diff \xc}
\newcommand*\laplace{\mathop{}\!\mathcal{4}}

\newcommand{\xr}{\mathbf{x}_r}
\newcommand{\xv}{\mathbf{x}_v}
\newcommand{\xw}{\mathbf{x}_w}
\newcommand{\xa}{\mathbf{x}_a}
\newcommand{\xb}{\mathbf{x}_b}
\newcommand{\xm}{\mathbf{x}_m}
\newcommand{\xmbar}{\overline{\mathbf{x}}_m}
\newcommand{\nonEmpty}{\Gamma}
\newcommand{\maskNonEmpty}{\textbf{m}_{\nonEmpty}}
\newcommand{\phasor}{\mathcal{P}}
\newcommand{\Pf}{\Fspace{\phasor}}
\newcommand{\pfIllumination}{\Fspace{I}}
\newcommand{\Pt}{{\phasor}}
\newcommand{\ptIllumination}{I}
\newcommand{\gating}{\mathbf{G}}
\newcommand{\dist}{\mathbf{D}}
\newcommand{\distf}[1]{\dist\left(#1\right)}
\newcommand{\gatingf}[1]{\gating\left(#1\right)}
\newcommand{\phasorf}[1]{\phasor\left(#1\right)}
\newcommand{\phasorwf}[1]{\phasor_{\pfFreq}\!\left(#1\right)}
\newcommand{\phasorxt}{\phasorf{\x, \pftime}}
\newcommand{\phasorwxt}{\phasorwf{\x, \pftime}}
\newcommand{\phasorxpt}{\phasorf{\xp, \pftime}}
\newcommand{\phasorwxpt}{\phasorwf{\xp, \pftime}}
\newcommand{\phasorxct}{\phasorf{\xc, \pftime}}
\newcommand{\phasorwxct}{\phasorwf{\xc, \pftime}}
\newcommand{\pfImagingModel}{\Phi}
\newcommand{\ROI}{V}
\newcommand{\pfImage}{I}
\newcommand{\pfImageof}[2]{\pfImage_{#1}\left(#2\right)}
\newcommand{\pfImpulse}{\stransportMatrixTransient}
\newcommand{\pfImpulseFun}{\pfImpulse\left(\xp,\xc,\pftime\right)}
\newcommand{\tof}{\textbf{\pftime}}
\newcommand{\dtof}[1]{\textbf{\pftime}_d\left(#1\right)}
\newcommand{\pfThinLens}{\mathcal{L}}
\newcommand{\pfThinLensFun}[2]{\pfThinLens_{#1}\!\left(#2\right)}
\newcommand{\pfFreq}{\omega}
\newcommand{\ff}{\Fspace{f}}
\newcommand{\fftc}{\Fspace{f}_{\textrm{tc}}}
\newcommand{\ffcc}{\Fspace{f}_{\textrm{cc}}}
\newcommand{\ffss}{\Fspace{f}_{\textrm{ss}}}
\newcommand{\ftc}{f_{\textrm{tc}}}
\newcommand{\fcc}{f_{\textrm{cc}}}
\newcommand{\fss}{f_{\textrm{ss}}}
\newcommand{\fbp}{f_{\textrm{bp}}}
\newcommand{\pathseq}[1]{\left\langle#1\right\rangle}

\newcommand{\lM}{M}
\newcommand{\lMp}{M^{\prime}} 
\newcommand{\lMG}{M^{G}} 
\newcommand{\lR}{R}
\newcommand{\lRp}{R^{\prime}}
\newcommand{\lG}{G}
\newcommand{\lGp}{G^{\prime}}
\newcommand{\lL}{\mathcal{L}}
\newcommand{\lS}{\mathcal{S}}
\newcommand{\lSp}{\mathcal{S}^{\prime}}
\newcommand{\lSpp}{\mathcal{S}^{\prime \prime}}
\newcommand{\lV}{\mathcal{V}}
\newcommand{\lW}{\mathcal{W}}
\newcommand{\xg}{\textbf{c}_\lG} 
\newcommand{\normalg}{\textbf{n}_\lG} 

\newcommand{\fM}{\droyo{REPLACE ME}}
\newcommand{\ffM}{\droyo{REPLACE ME}}
\newcommand{\ftcM}{f_{\textrm{tc\lM}}}
\newcommand{\fftcM}{\Fspace{f}_{tc\textrm{\lM}}}
\newcommand{\fccM}{f_{\textrm{cc\lM}}}
\newcommand{\ffccM}{\Fspace{f}_{cc\textrm{\lM}}}

\title{Virtual Mirrors: Non-Line-of-Sight Imaging Beyond the Third Bounce}


\author{Diego Royo}
\email{droyo@unizar.es}
\orcid{0000-0001-6880-322X}
\affiliation{%
   \institution{Universidad de Zaragoza -- I3A}
   \city{Zaragoza}
   \country{Spain}
}

\author{Talha Sultan}
\email{tsultan@wisc.edu}
\orcid{0000-0002-8372-2855}
\affiliation{%
   \institution{University of Wisconsin--Madison}
   \city{Madison, WI}
   \country{USA}
}

\author{Adolfo Muñoz}
\email{adolfo@unizar.es}
\orcid{0000-0002-8160-7159}
\affiliation{%
   \institution{Universidad de Zaragoza -- I3A}
   \city{Zaragoza}
   \country{Spain}
}

\author{Khadijeh Masumnia-Bisheh}
\email{khadijeh.masumnia-bisheh@wisc.edu}
\orcid{0000-0001-8464-9955}
\affiliation{%
   \institution{University of Wisconsin--Madison}
   \city{Madison, WI}
   \country{USA}
}

\author{Eric Brandt}
\email{elbrandt@wisc.edu}
\orcid{0000-0003-0206-182X}
\affiliation{%
   \institution{University of Wisconsin--Madison}
   \city{Madison, WI}
   \country{USA}
}

\author{Diego Gutierrez}
\email{diegog@unizar.es}
\orcid{0000-0002-7503-7022}
\affiliation{%
   \institution{Universidad de Zaragoza -- I3A}
   \city{Zaragoza}
   \country{Spain}
}

\author{Andreas Velten}
\email{velten@wisc.edu}
\orcid{0000-0001-5591-828X}
\affiliation{%
   \institution{University of Wisconsin--Madison}
   \city{Madison, WI}
   \country{USA}
}

\author{Julio Marco}
\email{juliom@unizar.es}
\orcid{0000-0001-9960-8945}
\affiliation{%
   \institution{Universidad de Zaragoza -- I3A}
   \city{Zaragoza}
   \country{Spain}
}

\renewcommand{\shortauthors}{Diego Royo, Talha Sultan, Adolfo Muñoz, Khadijeh Masumnia-Bisheh, Eric Brandt, Diego Gutierrez, Andreas Velten and Julio Marco}

\begin{abstract}
  Non-line-of-sight (NLOS) imaging methods are capable of reconstructing complex scenes that are not visible to an observer using indirect illumination. 
However, they assume only third-bounce illumination, so they are currently limited to single-corner configurations, and present limited visibility when imaging surfaces at certain orientations.
\new{To reason about and tackle these limitations, we make the key observation that planar diffuse surfaces behave specularly at wavelengths used in the computational wave-based NLOS imaging domain. We call such surfaces \emph{virtual mirrors}.
We leverage this observation to expand the capabilities of NLOS imaging using illumination beyond the third bounce, addressing two problems: imaging single-corner objects at limited visibility angles, and imaging objects hidden behind two corners.}
\new{To image objects at limited visibility angles,} we first analyze the reflections of the known illuminated point on surfaces of the scene as \NEWW{an} estimator of \NEWW{the position and orientation of objects with limited visibility}.
We then image those \new{limited} visibility objects by computationally building secondary apertures at other surfaces that observe the target object from a \new{direct} visibility perspective. 
Beyond single-corner NLOS imaging, \new{we exploit the specular behavior of {virtual mirrors}} to image objects hidden behind a second corner \new{by imaging the space behind such virtual mirrors}, where the mirror image of objects hidden around two corners is formed.  
No specular surfaces were involved in the making of this paper.  
\end{abstract}





\maketitle

\section{Introduction}
\label{sec:introduction}

\begin{figure}
    \centering
    \def\svgwidth{0.92\columnwidth}
    \captionsetup{skip=9pt}
    \begin{small}
\begingroup%
  \makeatletter%
  \providecommand\color[2][]{%
    \errmessage{(Inkscape) Color is used for the text in Inkscape, but the package 'color.sty' is not loaded}%
    \renewcommand\color[2][]{}%
  }%
  \providecommand\transparent[1]{%
    \errmessage{(Inkscape) Transparency is used (non-zero) for the text in Inkscape, but the package 'transparent.sty' is not loaded}%
    \renewcommand\transparent[1]{}%
  }%
  \providecommand\rotatebox[2]{#2}%
  \newcommand*\fsize{\dimexpr\f@size pt\relax}%
  \newcommand*\lineheight[1]{\fontsize{\fsize}{#1\fsize}\selectfont}%
  \ifx\svgwidth\undefined%
    \setlength{\unitlength}{1586.53149222bp}%
    \ifx\svgscale\undefined%
      \relax%
    \else%
      \setlength{\unitlength}{\unitlength * \real{\svgscale}}%
    \fi%
  \else%
    \setlength{\unitlength}{\svgwidth}%
  \fi%
  \global\let\svgwidth\undefined%
  \global\let\svgscale\undefined%
  \makeatother%
  \begin{picture}(1,0.51693902)%
    \lineheight{1}%
    \setlength\tabcolsep{0pt}%
    \put(0,0){\includegraphics[width=\unitlength,page=1]{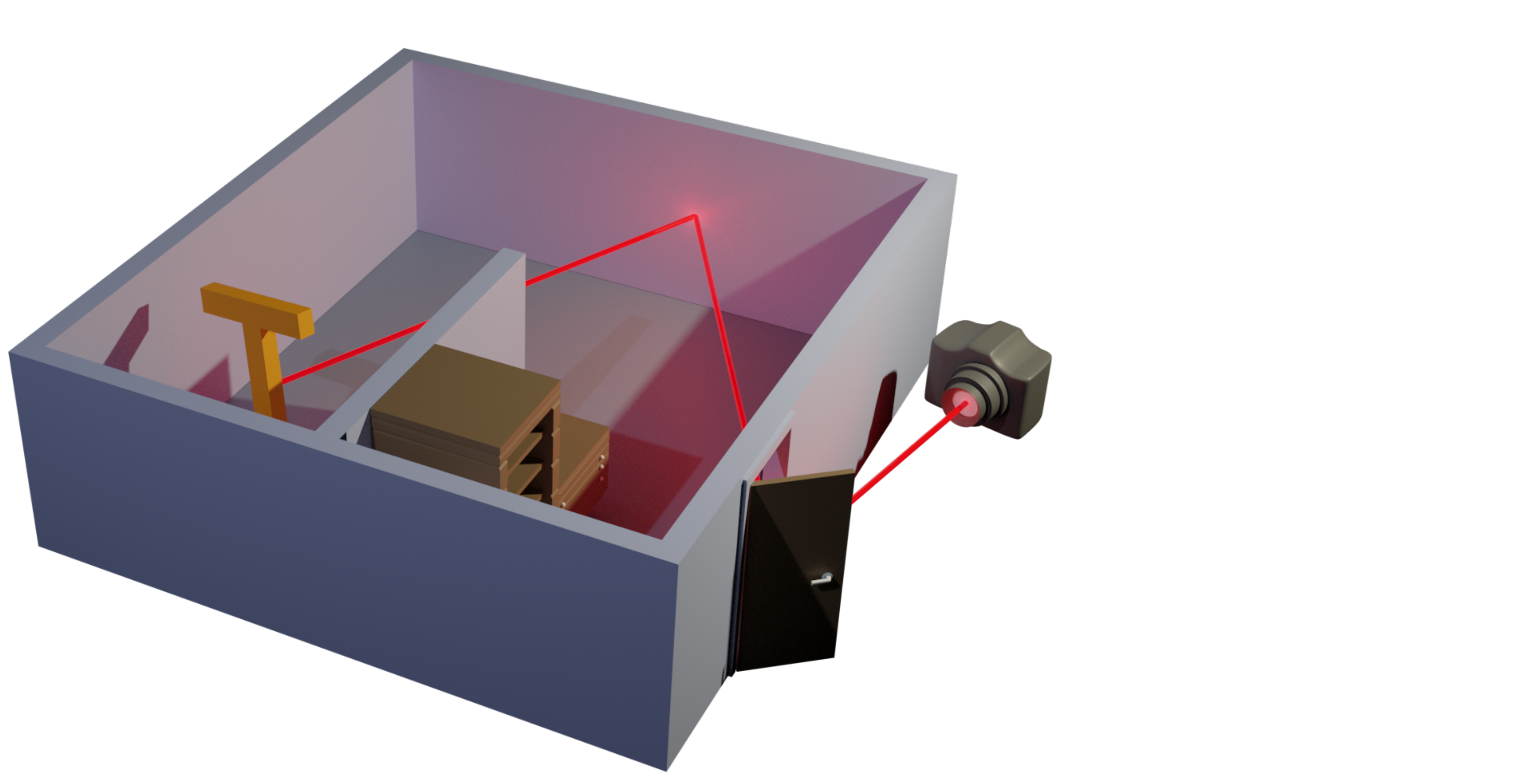}}%
    \put(0.86997195,0.41113435){\color[rgb]{0,0,0}\makebox(0,0)[t]{\lineheight{1.25}\smash{\begin{tabular}[t]{c}Computed image of\\T-shaped object\\from a real capture\end{tabular}}}}%
    \put(0.6125765,0.00453817){\color[rgb]{0,0,0}\makebox(0,0)[t]{\lineheight{1.25}\smash{\begin{tabular}[t]{c}Relay surface\end{tabular}}}}%
    \put(0.58421437,0.49848662){\color[rgb]{0,0,0}\makebox(0,0)[t]{\lineheight{1.25}\smash{\begin{tabular}[t]{c}Diffuse surface\\(\emph{Virtual mirror})\end{tabular}}}}%
    \put(0,0){\includegraphics[width=\unitlength,page=2]{teaser-v11.pdf}}%
  \end{picture}%
\endgroup%

    \end{small}
    \caption{\new{We image geometry around two corners. The displayed room only has diffuse surfaces. A single laser source illuminates the relay surface, then the resulting indirect illumination is captured and used as input to the problem. The main observation in our work shows that, by only applying computations to the captured data, surfaces that are diffuse in the real world can exhibit specular properties, based on well-known wave interference principles. We leverage this observation in the room example shown,} using one of the diffuse surfaces as a \emph{virtual mirror} to image a T-shaped geometry, that would otherwise not be visible by only looking around one corner.}
    \vspace{-1em}
    \label{fig:teaser}
    \Description[We image a T-shaped geometry around two corners]{The virtual mirror surface is oriented such that you would see the T-shaped geometry from the first corner, if the virtual mirror was an actual mirror.}
\end{figure}

Non-line-of-sight (NLOS) imaging methods retrieve information of scenes that are hidden from the observer, including geometric reconstructions \cite{Velten2012nc,wu2021non}, position detection \cite{bouman2017turning,Yi2021SIGA}, or motion tracking \cite{gariepy2016detection}, with many applications in fields such as remote sensing, autonomous driving or biological imaging. 
Under this regime, methods that image scenes hidden around a corner have shown promising results thanks to ultra-fast imaging devices (e.g., \cite{velten2013femto,buttafava2015non,shin2016photon}). Such time-gated NLOS imaging methods provide detailed reconstructions of hidden scenes by triangulating geometric positions using the time of flight of round-trip third-bounce illumination paths between a visible relay surface and the hidden scene \cite{Velten2012nc,OToole2018confocal,Xin2019theory,Lindell2019wave}.

These methods operate under the assumption of third-bounce-only illumination, with higher-order illumination usually degrading the reconstructions due to ambiguities in the time of flight of light. Recent wave-based NLOS imaging methods have shown how such higher-order bounces can be isolated from third-bounce illumination and visualized, based on an imaging paradigm that interprets the time-resolved illumination captured at a relay surface as light arriving at a virtual aperture \cite{Liu2020phasor,Liu2019phasor,nam2021low,Marco2021NLOSvLTM}, effectively transforming \NEWW{the relay surface} into a virtual line-of-sight (LOS) imaging system.

In our work, we demonstrate how such higher-order bounces can be used to expand the capabilities of existing NLOS imaging systems, and overcome some of its current limitations.
In particular, we draw a parallelism between Huygens' principle and the recent wave-based phasor-field NLOS imaging formulation \cite{Liu2019phasor}.
We intuitively show how, due to well-known wave interference principles, \new{surfaces that are diffuse under visible light can} behave like mirrors during the \new{computational} NLOS \new{wave-based} imaging process; we call such surfaces \textit{virtual mirrors}. \new{It is thus important to understand that virtual mirrors only show specular behavior in the computational domain, since real NLOS capture systems still receive diffuse illumination.}

From this observation, we show how (in the computational domain) specular reflections in the form of \textit{fourth-} and \textit{fifth-bounce} illumination actually encode useful information about the scene, then leverage these higher-order bounces to address two longstanding problems in NLOS imaging: visibility issues due to the missing-cone problem, and looking around \textit{two} corners.

The missing-cone problem is inherent to all third-bounce NLOS imaging methods, and refers to the fact that \new{certain scene configurations and features cannot be accessed by NLOS measurements depending on their position and orientation with respect to the relay surface  \cite{Liu2019analysis}. }\new{The corresponding surfaces are said to be inside the null-reconstruction space of third-bounce methods.}

To image surfaces inside such null-reconstruction space, we first image the mirror reflections of a known illuminated point \neww{on the relay surface, produced by all surfaces of the scene, including those \new{surfaces inside the null-reconstruction space}. By analyzing these reflections, we} infer the position and orientation of the hidden surfaces \NEW{inside the null-reconstruction space} that produced such mirror reflections. 
\newww{To avoid ambiguities introduced by inference, we introduce a novel procedure} to directly image the hidden surfaces by \neww{creating} a \emph{second} virtual aperture at other scene surfaces.

In addition, we propose a second novel \neww{procedure to} image objects hidden behind \emph{two} corners (\fref{fig:teaser}). 
For this, we show how to use fifth-bounce illumination to image the space behind a diffuse hidden surface, which effectively acts as a mirror as seen from the relay surface, \newww{allowing us to observe a mirror image of the object hidden behind two corners}. 
\neww{Our key insight for this second procedure is selecting the location of the volume being imaged, which is in principle orthogonal to the particular imaging algorithm used.} Given our virtual mirror surfaces, we target the reflected space \emph{behind} such surfaces, where mirror images are formed (just like with real mirrors). \newww{We demonstrate our procedure} works even if the intermediate surface itself falls \new{inside the null-reconstruction space}. 

\neww{In summary, we demonstrate how to extend the capabilities of existing NLOS imaging methods, by i) imaging surfaces inside the null-reconstruction space by leveraging fourth-bounce illumination, and ii) imaging surfaces hidden around two corners using fifth-bounce illumination. }
\neww{We validate our findings both in simulation and using real captured data. Last, all of our experiment data, simulation and imaging software are publicly available\footnote{\href{https://graphics.unizar.es/projects/VirtualMirrors_2023}{https://graphics.unizar.es/projects/VirtualMirrors\_2023}}.}

\section{Related work}
\label{sec:related-work}

\new{NLOS imaging methods \neww{analyze indirect illumination from paths that scatter one or multiple times in the target hidden scene to obtain information about it}. They can be divided into active and passive methods. Active methods \neww{use controlled light sources to illuminate} the hidden scene~\cite{velten2013femto, katz2014non, cao2022high, luesia2022non}, while passive methods rely on ambient illumination~\cite{bouman2017turning, krska2022double} \neww{or light emitted by hidden objects themselves}~\cite{saunders2019computational}. In our work, an active laser source emits light pulses to illuminate the hidden scene.}

\paragraph{Time-gated NLOS imaging}
The ability to capture time-gated illumination using time-of-flight detectors at picosecond resolution has nurtured a wide range of NLOS imaging methods ~\cite{Faccio2020non,maeda2019recent,Jarabo2017transient,pediredla2019snlos}. The first methods to demonstrate high-quality 3D reconstructions employed ellipsoidal backprojection \cite{Velten2012, buttafava2015non} acquired for non-confocal optical paths.
These methods operate on the time domain of the captured light transport and are computationally expensive. 
By restricting data acquisition to confocal optical light paths, NLOS reconstructions can be formulated as a closed-form, deconvolution-based linear inverse problem which can be solved efficiently in the frequency domain \cite{Lindell2019wave, OToole2018confocal}. Since they are based on frequency-space deconvolutions, they require the use of a regular sampling grid on a planar relay \neww{surface}. Furthermore, in contrast to our work, both ellipsoidal-based and deconvolution-based techniques are unable to account for light bounces beyond the third.

\paragraph{Phasor-field formulation}
The phasor-field \neww{formulation \cite{Liu2019phasor} provides wave-based models to propagate} virtual waves into the hidden scene, which \neww{allows to turn} a visible relay surface into a virtual LOS camera. This allows to build fundamental wave-optics parallelisms between forward operators in LOS imaging and the backprojection operators that drive time-gated NLOS imaging approaches~\cite{rezaPhasorFieldWaves2019, rezaPhasorFieldWaves2019a, doveParaxialPhasorfieldPhysical2020, dove2020nonparaxial, dove2020speckled, teichmanPhasorFieldWaves2019, Guillen2020Effect, laurenzis2022time}. 
The phasor-field formalism allows to image hidden scenes using either confocal or non-confocal setups~\cite{Liu2019phasor}. Subsequent implementations have gained efficiency by working in the frequency domain \cite{Liu2020phasor}, which has led to interactive and real-time reconstructions of dynamic hidden scenes~\cite{nam2020real,liao2021fpga} at the cost of using a regular sampling grid. Moreover, it allows to use both planar and non-planar relay surfaces~\cite{manna_non-line--sight-imaging_2020}, and to leverage known occlusions in the reconstructions ~\cite{dove2019paraxial}. For memory-constrained applications, different propagation operators can be implemented, such as zone plates \cite{luesia2023zone}. 
Last, \citet{Marco2021NLOSvLTM} \NEW{leveraged the phasor-field formulation} to separate direct and indirect illumination of hidden scenes by combining exhaustive scans of both laser and sensor positions. In this work, we build on top of the virtual-wave LOS parallelisms and reason about the virtual reflectivity of hidden surfaces. We then leverage higher-order illumination to image geometry \neww{hidden around two corners}, as well as to \neww{directly} estimate hidden objects that \neww{have limited visibility to} classic third-bounce methods.

\paragraph{NLOS with specular reflections} 
Prior works utilized \new{actual} specular reflections by using centimeter-scale acoustic waves for NLOS reconstructions \cite{lindell2019acoustic}, or millimeter-scale radio waves for tracking hidden objects \cite{scheiner_seeing_2020}. Using specular reflections requires directly sampling the path from the hidden scene to the relay surface, needing strong assumptions about surface albedo and orientation~\cite{lindell2019acoustic}, or placing the scanning system far from the relay surface~\cite{scheiner_seeing_2020}. 
Specular reflections produced by infrared wavelenghts have also been utilized for \neww{passive} NLOS~\cite{maeda2019thermal, kaga_thermal_2019}.
These systems are also restricted to reconstructions of planar scenes or to object tracking inside the hidden scene. In our work we observe that, when computationally transforming the temporal profile of diffuse surfaces into the frequency domain (following the phasor-field model~\cite{Liu2019phasor}), they exhibit mirror-like reflectance properties. We leverage higher-order illumination bounces produced by these virtual mirrors to overcome classical NLOS visibility challenges and to look around an additional second corner.

\section{Background and insights}
\label{sec:background}

\new{In the following, we cover the background on wave-based NLOS imaging that forms the basis of our work, as well as related insights to understand our contributions. 
\NEWW{In \sref{sec:nlos-imaging-limitations}} we describe the two key existing challenges that we address in this paper.}
\subsection{Wave-based NLOS image formation}
\begin{figure}
    \centering
    \captionsetup{skip=3pt}
    \def\svgwidth{\columnwidth} 
    \begin{small}
    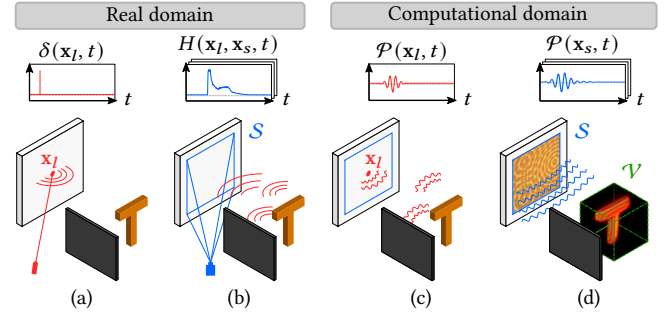
    \end{small}
    \caption{
    (a) A \NEW{laser device emits a} delta light pulse $\delta(\xl,t)$ \NEW{that} illuminates a point $\xl$ on the relay surface.
    (b) An ultra-fast \NEW{sensing device} captures time-resolved illumination $H(\xl,\xs,t)$ on multiple visible points $\xs$ of the imaging aperture $\lS$.
    (c) By convolving the impulse response $H(\xl, \xs, t)$ with an illumination function $\Pt(\xl, t)$, the phasor-field framework obtains the response of the \NEW{hidden} scene $\Pt(\xs, t)$ to any arbitrary illumination function using only computation.
    (d) Lastly, the relay surface acts as a computational lens that focuses (propagates \new{and adds}) \NEW{the response} $\Pt(\xs, t)$ from all points $\xs$ at \neww{each point $\xv$ in the bounding volume $\lV$ of the hidden scene,} effectively transforming the NLOS problem into a virtual LOS problem. For this example, we \neww{compute the time-resolved image $\fcc(\xv, t)$ using the confocal camera model which, when evaluated at $t=0$, shows the T-shaped object.}}
    \label{fig:RSD_image_formation}
    \vspace{-2.5em}
\end{figure}

\new{In a classic NLOS setup, the capture device illuminates and measures indirect light at a diffuse relay surface, lacking direct line of sight to the target scenes being imaged. The phasor-field formulation \cite{Liu2019phasor} brings the NLOS problem into a \textit{virtual} LOS domain, creating computational imaging devices at the relay surface which can directly illuminate and \NEWW{capture} the hidden scene.}
To understand our work, it is key to distinguish between the real and computational imaging domains in the phasor-field formulation.
First, in the real domain, the acquisition process involves a laser \NEW{device} that emits illumination pulses towards locations $\xl$ on the relay surface (\fref{fig:RSD_image_formation}a). The resulting indirect illumination produced by the hidden scene is then captured by an ultra-fast \NEW{sensing device} at points $\xs\in\lS$ on the relay surface (\fref{fig:RSD_image_formation}b), yielding a time-resolved impulse response function $H(\xl, \xs, t)$, where $t$ represents time.
\new{In the second stage, the phasor-field framework operates on $H$ to \emph{computationally} illuminate the hidden scene (\fref{fig:RSD_image_formation}c) and then \newww{compute} images \newww{of the hidden scene under such illumination} (\fref{fig:RSD_image_formation}d). Note that, while the first stage illuminates and senses the scene in the real domain, the second stage (which we describe in the rest of this section) is entirely computational.}

\newww{The time of flight between the laser device and $\xl$ (\fref{fig:RSD_image_formation}a, red) and between $\xs \in \lS$ and the sensing device (\fref{fig:RSD_image_formation}b, blue) introduces temporal delays on the illumination captured in $H(\xl,\xs,t)$. In practice, we shift the temporal dimension of $H$ 
so that $t$ represents the time of flight of light paths that start at $\xl$, scatter in the hidden scene, and end at $\xs$. Placing the origin of the time reference system at the relay surface instead of at the laser and sensing devices is common practice in NLOS imaging and does not affect the algorithms \cite{Liu2019phasor,Marco2021NLOSvLTM}.}

\new{\paragraph{Computationally illuminating the hidden scene}
Points $\xl$ in the relay surface reflect delta illumination pulses emitted by the laser. \neww{In \NEW{the} computational domain, these points $\xl$ are now the \emph{emitters}.}
Given the impulse response function $H(\xl, \xs, t)$, captured from a delta illumination pulse $\delta(\xl, t)$, we can compute the response of the scene to any other arbitrary time-resolved illumination function $\Pt(\xl, t)$ emitted from $\xl$ (\fref{fig:RSD_image_formation}c). The resulting time-resolved response $\Pt(\xs, t)$ at points $\xs$ on the relay surface is computed as
\begin{align}
    \Pt(\xs, t) = \int\limits_{\mathcal{L}} \Pt(\xl, t) \ast H(\xl, \xs, t) \diff \xl,
    \label{eq:virtual_illumination}
\end{align}
where $\ast$ represents a convolution in time. Throughout \neww{our} paper, we only use a single point $\xl$ located at the center of the relay surface.
Following the work by \citet{Liu2019phasor}, for the illumination function $\Pt(\xl, t)$  we use a pulse wave with a Gaussian envelope (\fref{fig:RSD_image_formation}c) with wavelength $\lambda_c$ and standard deviation $\sigma$:
\begin{align}
    \Pt(\xl, t) = e^{i 2 \pi \frac{t}{\wl_c} - \frac{1}{2} \left( \frac{t}{\sigma} \right)^2 }.
    \label{eq:gaussian-pulse}
\end{align}
We discuss the choice of $\wl_c$ and $\sigma$ and their implications in the imaging process in \sref{sec:light_interference}.
}

\paragraph{The \NEW{virtual} camera analogy}
The phasor-field framework treats the relay surface as the computational lens in a virtual camera that directly observes the hidden scene, with an aperture $\lS$ defined by points $\xs \in \lS$. 
The lens operators used in the phasor-field framework are well-known wave-based lens imaging operators \cite{Liu2019phasor,Liu2020phasor} defined as a function of each frequency component $\fq$ of the signal.
Because of this, we transform $\Pt(\xs, t)$ to the frequency domain by applying a Fourier transform over the time domain, obtaining the complex-valued field $\Pf(\xs, \fq) = \mathcal{F}\left\{ \Pt(\xs, t) \right\}$.
Each complex value $\Pf(\xs, \fq)$ represents a wave (i.e., the values form a phasor field) resulting from illuminating the scene with the phasor $\Pf(\xl, \fq) = \mathcal{F}\left\{ \Pt(\xl, t) \right\}$, with illumination frequency $\fq$. In the camera analogy, $\Pf(\xs, \fq)$ can be understood as out-of-focus illumination from the hidden scene, and the goal of the computational lens is to focus the phasors $\Pf(\xs, \fq)$ to form an in-focus image of the hidden scene (\fref{fig:RSD_image_formation}d).
This focusing operation is specific to the chosen imaging operator $\Phi$, which defines the properties of the \NEW{virtual} camera:
\begin{align}
    \Fspace{f}(\xv, \fq) = \Phi(\xv, \Pf(\xs, \fq)),
    \label{eq:imaging_operator}
\end{align}
where $\Fspace{f}(\xv, \fq)$ is the resulting image of the hidden scene computed for an imaging frequency $\fq$ at points $\xv$ on the \newww{focal} plane \NEW{of the virtual camera.}
\newww{Applying the focusing operation from the phasor field $\Pf(\xs, \fq)$ at a point on a plane behind the relay surface is equivalent to focusing at symmetric locations in front of the relay surface.}
For clarity in the explanations, we use the latter approach in the rest of the paper, 
\NEW{where $\xv$ denotes} a point \NEWW{in} the bounding volume $\lV$ of the hidden scene (\fref{fig:RSD_image_formation}d).
\new{
\newww{Due to the large size of the aperture $\lS$ at the relay surface, imaging the hidden scene with a single focal plane (as in conventional photography) results in a very shallow depth of field, which yields out-of-focus illumination from objects outside \NEW{the focal plane} \cite{Marco2021NLOSvLTM}. To mitigate this problem, we use the computational lens 
\NEW{to sweep the focal plane across the hidden scene, capturing} a sequence of planar images taken at several focal distances \NEW{(i.e., creating a focal stack). We} arrange these planar images to form a volumetric image of the \NEW{scene contained in} $\lV$.}}

\new{\paragraph{Lens imaging operators.} 
The analogy of a computational lens focusing at any point $\xv$ \NEW{in $\lV$} is defined, in practice, by the propagation of the phasors \neww{$\Pf(\xs, \fq)$} from all points $\xs$ in the aperture $\lS$ to $\xv$.
In general, when light travels from any point $\xa$ to another point $\xb$, the phasor $\Pf(\xa, \fq)$ at $\xa$ undergoes a phase shift and attenuation modeled by the Rayleigh-Sommerfeld Diffraction (RSD) operator. The phasor at $\xb$, $\Pf(\xb, \fq)$, is then computed as
\begin{align}
    \Pf(\xb, \fq) = \Pf(\xa, \fq) \frac{e^{i k \norm{\xb - \xa}}}{\norm{\xb - \xa}},
    \label{eq:phase_shift}
\end{align}
where $\norm{\xb - \xa}$ is the optical distance between $\xa$ and $\xb$. The wavenumber $k = 2\pi\fq/c$ (where $c$ is the speed of light) is the conversion factor from optical distance to phase. The phase shift (numerator) and attenuation (denominator) form the RSD operator.}
\new{ 
The phasor-field formulation uses RSD operators to define its imaging operators $\Phi$, and \NEWW{capture} images $\Fspace{f}(\xv, \fq)$ of the hidden scene with different characteristics. These images can be defined under the following general expression:
\begin{align}
    \Fspace{f}(\xv, \fq) = \int \limits_{\lS} \frac{e^{i k t_s c}}{t_s c} \int \limits_{\lL} \frac{e^{i k t_l c}}{t_l c}  \Pf(\xl, \fq) \Fspace{H}(\xl, \xs, \fq) \diff \xl \diff \xs,
    \label{eq:RSD_freq}
\end{align}
\neww{where $t_s$ and $t_l$ are parameters of the RSD operators, representing time of flight, and are used to implement the two different camera models \NEW{in our work, as} \NEWW{explained in \sref{sec:time-resolved-models}}. 
Note that the imaging frequency $\fq$ is included in the wavenumber $k=2\pi\fq / c$ of these operators.}
$\Pf(\xl, \fq)$ and $\Fspace{H}(\xl,\xs,\fq)$ are the frequency-domain counterparts of $\Pt(\xl, t)$ and $H(\xl,\xs,t)$, respectively, obtained via Fourier transform.
The innermost integral over $\lL$ is the frequency-domain counterpart of the illumination function (\eref{eq:virtual_illumination}), but including an RSD operator.
This allows for more general imaging operators by interpreting $\lL$ as an illumination aperture where another lens \neww{on the relay surface can focus the emitted illumination at specific locations $\xv$ in the scene. In our case, with only a single point $\xl$ in $\lL$, this focus operation simply applies a phase shift and attenuation to the illumination phasor $\Pf(\xl, \fq)$\newww{; note that we can apply a different focus \NEW{ operation} at each location $\xv$ in the scene.} In the time domain, this phase shift effectively shifts the time instant at which each point $\xv$ in the scene is computationally illuminated using $\Pt(\xl, t)$.
In the following, we summarize the two camera models used in our work, which result from choosing specific values for the parameters $t_s$ and $t_l$ in \eref{eq:RSD_freq}.}}

\subsection{\new{Time-resolved camera models}} 
\label{sec:time-resolved-models}
\new{Throughout our paper we implement two \neww{camera} models introduced by previous work \cite{Liu2019phasor}: the \textit{transient} camera model and the \textit{confocal} camera model.}
These two models allow \NEW{us} to address different challenges in NLOS imaging in this work, such as the missing-cone problem and imaging objects hidden around two corners, by analyzing \newww{captured images of the scene} \neww{at different locations and time instants.}
\newww{These models compute time-resolved images of the hidden scene via inverse Fourier transform over the frequency domain.
We use $\ftc(\xv,t)=\mathcal{F}^{-1}\{\fftc(\xv, \fq)\}$ and $\fcc(\xv,t)=\mathcal{F}^{-1}\{\ffcc(\xv, \fq)\}$ to refer to time-resolved images computed using the transient and confocal camera models \NEW{$\ftc(\xv,t)$ and $\fcc(\xv,t)$}, respectively. \NEWW{These models result from choosing specific values of $t_s$ and $t_l$ in \eref{eq:RSD_freq}}.}

\newww{
As previously discussed, we mitigate \NEW{depth of field} issues coming from the large camera aperture by computing multiple planar images at different focal distances \NEW{that cover the volume $\lV$}, which form a focal stack\NEWW{; we arrange these planar images} to form a volumetric image $\Fspace{f}(\xv, \Omega)$ \NEW{for all points $\xv$ of the hidden scene}.
Any hidden scene element \NEW{located at $\xv$} will therefore be in focus in one planar image \NEW{that forms $\Fspace{f}(\xv, \Omega)$.}
Due to the time-frequency correspondence, the resulting time-resolved volumetric image $f(\xv,t) = \mathcal{F}^{-1}\{\Fspace{f}(\xv, \fq)\}$ is composed of a set of time-resolved planar images in the focal stack.
\NEWW{In conventional photography, a focal stack contains multiple \NEWW{planar} images of the same scene captured at different focal distances.}
The case of $f(\xv,t)$ is analogous, \NEWW{but} also \NEWW{including} the temporal dimension: \NEW{each frame at $t$ of $f(\xv, t)$}
combines multiple planar images \NEWW{that capture the hidden scene at a different focal distances and at different time instants}.
\NEWW{Due to this effect}, light transport events that occurred at the same time \NEWW{in the scene} are captured in a different order \NEWW{on each planar image that forms the time-resolved volumetric image $f(\xv, t)$}.
Consequently, each frame of $f(\xv,t)$ may simultaneously show multiple light transport events of the hidden scene, despite these occurred at different time instants.
In our work we compute time-resolved volumetric images $\ftc(\xv,t)$ and $\fcc(\xv,t)$ to address different challenges of NLOS imaging by analyzing light transport events in the hidden scene.
\NEWW{These events have occurred at different instants, but each frame $t$ captures them simultaneously.}}

\paragraph{Transient camera. }
\new{The transient camera model implements a \NEW{computational lens located at} $\lS$, focused at hidden scene points $\xv$ with $t_s = \norm{\xs-\xv}/c$, \newww{and it does not implement any lens for the illumination aperture $\lL$. For this, the RSD operator in $\lL$ of \eref{eq:RSD_freq} is ignored, resulting in}}
\begin{align}
    \fftc(\xv, \fq) = \int \limits_{\lS} \frac{e^{i k \norm{\xs - \xv}}}{\norm{\xs - \xv}} \int \limits_{\lL} \Pf(\xl, \fq) \Fspace{H}(\xl, \xs, \fq) \diff \xl  \diff \xs.
    \label{eq:RSD_freq_tc}
\end{align}
\neww{The computed images $\ftc(\xv,t) = \invFourier{\fftc(\xv,\fq)}$ of this \neww{camera} model resemble the captures of existing time-resolved cameras \cite{velten2013femto,Heide2013} if the hidden scene were illuminated using $\Pt(\xl, t)$.}
\newww{However, as previously discussed, the computed images $\ftc(\xv,t)$ capture the hidden scene at a different time
for every point $\xv$. 
\NEWW{As a result from the focusing operation in \eref{eq:RSD_freq_tc}}, the hidden scene elements that are in focus at $\xv$ will be captured in $\ftc(\xv, t)$ in the frame with time $t$ corresponding to the time of flight between $\xl$ and the location of the \NEW{\emph{actual}} scene element \NEWW{captured at $\xv$}. Importantly, as we deal with mirror reflections in this work, note that the location of the \NEWW{actual} scene element may not correspond with $\xv$ \NEWW{if $\ftc(\xv,t)$ captures} a mirror reflection of such element at $\xv$. \NEWW{For actual elements of the hidden scene located at $\xv$, the computed images are a good approximation of time-resolved light transport at such elements under computational illumination.}}
\neww{In our work, we use this camera model to analyze mirror reflections of known elements (e.g., the known illuminated point $\xl$) in the hidden scene, and show how to leverage them to 
\NEW{address the missing-cone problem.}}

\paragraph{Confocal camera. }\new{We also implement a confocal camera model, which creates two \neww{computational lenses located at} $\lL$ and $\lS$, focused at the same hidden scene point $\xv$\NEW{, respectively} using $t_l = \norm{\xv-\xl}/c$ and $t_s = \norm{\xs-\xv}/c$, resulting in}
\begin{align}
    \hspace{-0.08em}\ffcc(\xv, \fq) = \!\! \int \limits_{\lS}\!\! \tfrac{e^{i k \norm{\xs - \xv}}}{\norm{\xs - \xv}}\!\! \int \limits_{\lL} \!\! \tfrac{e^{i k \norm{\xv - \xl}}}{\norm{\xv - \xl}} \Pf(\xl, \fq) \Fspace{H}(\xl, \xs, \fq) \diff \xl  \diff \xs.
    \label{eq:RSD_freq_cc}
\end{align}
\new{\neww{Throughout our work, we use a single \NEW{illuminated} point $\xl$, which yields a special case of $\fcc(\xv,t) = \invFourier{\ffcc(\xv, \fq)}$ \NEW{for the frame at} $t=0$.
The phase shift defined by $t_l = \norm{\xv-\xl}/c$ in the illumination aperture $\lL$ is equivalent to a temporal shift of the illumination function $\Pt(\xl,t)$
\newww{\NEW{corresponding to the} time of flight from $\xl$ to \NEW{each} location $\xv$. 
Similar to the transient camera model, \NEWW{the computed images $\fcc(\xv,t)$ capture the hidden scene at a different time for every imaged point $\xv$}}.}
\NEW{For this camera model, the frame at \mbox{$t=0$} of} $\fcc(\xv, t)$ represents direct illumination from the emitter \newww{at scene elements that are in focus at} $\xv$.
This is equivalent to the imaging models used by the vast majority of existing time-gated NLOS imaging methods \cite{Velten2012nc, OToole2018confocal, Lindell2019wave, Xin2019theory, ahn2019convolutional,Liu2019phasor} to reconstruct single-corner hidden scenes, which consider light with the shortest path $\xl \rarr \xv \rarr \xs$ a good estimator of the hidden geometry at $\xv$. 
In our work, we use this camera model as an intermediate step to address the missing-cone problem, and we show how to leverage it to image objects hidden behind \textit{two} corners.}

\section{Current NLOS imaging limitations}
\label{sec:nlos-imaging-limitations}
In this paper we address two of the fundamental limitations of NLOS imaging: the missing-cone problem and single-corner imaging, which we summarize in the following.

\subsection{\neww{The missing-cone problem}}
\label{sec:challenges:missing-cone}
\new{The missing cone is a fundamental problem of NLOS imaging where, for a relay surface of a given size, certain hidden surfaces cannot be accessed by NLOS measurements (i.e., the impulse response $H$), and thus cannot be reconstructed regardless of the imaging method employed. This problem is inherent to other well-established imaging methods as well e.g., computed tomography \cite{benning2015tgv,delaney1998globally}, which assume \NEW{three}-bounce transport.}
\new{We use the term \emph{null-reconstruction space} to denote the set of such surfaces that cannot be accessed by NLOS measurements. Note that surfaces inside the null-reconstruction space may still reflect some light towards the relay surface. However, their response $H$ only changes its unmodulated components (frequency $\fq = 0$), lacking the required modulated changes (components with frequency $\fq > 0$) to be reconstructed using third-bounce time-of-flight information. }
\new{The in-depth analysis by \citet{Liu2019analysis} shows that the missing-cone problem is universal across any NLOS measurement, and therefore affects both the transient camera $\ftc$ and the confocal camera $\fcc$ models.}
\new{In Section \ref{sec:contribution-2:goal-2}, we intuitively address the missing-cone problem from our virtual mirror analogy, and use fourth-bounce illumination to image objects that are inside the null-reconstruction space of third-bounce methods.}

\subsection{\neww{Single-corner imaging}}
\label{sec:challanges:single-corner}
\new{The existing time-resolved NLOS imaging methods are limited to reconstructing objects hidden behind a single corner (\fref{fig:RSD_image_formation}), based on third-bounce illumination assumptions. However, general scenes typically contain objects hidden behind several occluders, creating higher-order illumination at the relay surface that is equally captured on the impulse response function $H$ by time-resolved sensors\neww{, degrading the imaging results of third-bounce methods}. In our work, we show how to leverage such higher-order illumination \neww{with applications such as imaging objects} hidden behind two corners in \sref{sec:contribution-2:goal-1}.}
 
\section{Diffuse surfaces as virtual mirrors}
\label{sec:light_interference}
A key observation in our work is that surfaces that are diffuse under visible light may still exhibit specular properties in \new{the computational NLOS wave imaging domain}. 
As we will show, leveraging this observation allows us to extend the current range of NLOS imaging capabilities, including imaging scenes that are hidden behind \textit{two} corners. 

In wave-based methods, time-resolved transport $\phasor(\xbf,t)$ at \new{any point $\xbf$} in the hidden scene becomes a phasor $\Pf(\xbf, \fq)$ in the frequency domain.
According to Huygens' principle, when reaching a surface $\lM$, this spherical wavefront will in turn generate multiple secondary spherical wavefronts. \new{As an example, consider a point light at $\xl$ \NEW{whose emission is} defined by a phasor $\Pf(\xl,\fq)$ that illuminates points $\xm$ on a planar surface $\lM$, resulting in phasors $\Pf(\xm,\fq)$. We can then} compute the resulting phasor at any point $\xv$ in a volume $\lV$ as a superposition of phasors \NEW{from $\xm$} by extending \eref{eq:phase_shift} as
\begin{align}
\begin{split}
    \Pf(\xv, \fq) & = \int_\lM \Pf(\xm, \fq) \frac{e^{i k \norm{\xv - \xm}}}{\norm{\xv - \xm}} \diff \xm \\
    & = \int_\lM \Pf(\xl, \fq) \frac{e^{i k \left(\norm{\xv - \xm} + \norm{\xm - \xl}\right)}}{\norm{\xv - \xm} \norm{\xm - \xl}} \diff \xm.
    \label{eq:phasor_single_scattering}
\end{split}
\end{align}
For diffuse surfaces which are planar with respect to the illumination wavelength $\wl = \Omega^{-1}$, the newly generated phasors \NEW{$\Pf(\xv, \fq)$} result in a specular reflection of the incoming wavefront \NEW{from $\Pf(\xl, \fq)$}. In practice, this means that while the surface \NEW{$\lM$} may reflect visible light in all directions, the transient modulations \neww{(components with frequency $\fq > 0$)} that we need to \NEW{image} the scene propagate in the \textit{specular} direction of the reflected \new{computational} wave.

\begin{figure}[t]
    \centering
    \captionsetup{skip=0pt}
    \def\svgwidth{\columnwidth} 
    \begin{small}
    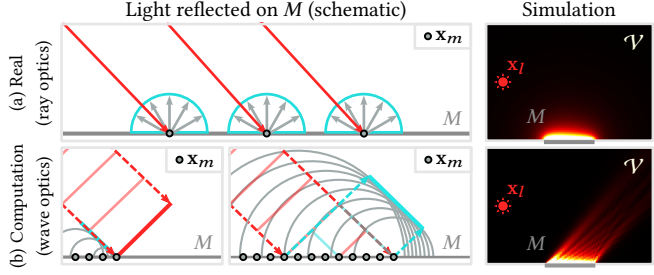
    \end{small}
    \caption{\new{Specular behavior of a diffuse surface in the computational wave-based domain.}
        (a) During the capture process, \neww{each point $\xm$ on a} diffuse surface $\lM$ reflects incoming light (red) isotropically (reflected rays in \new{grey}, reflection shape in \new{cyan}). On the right, the light from $\xl$ reflected on \neww{all points $\xm$ in the} surface $\lM$ is simulated \neww{for each point $\xv$} in $\lV$ using \new{ray} optics and Monte Carlo and the steady-state time average is shown. In the Fourier domain this is the component \NEW{with frequency} $\fq=0$. (b) When considering only \NEW{modulated light (i.e., choosing a Fourier component with frequency $\fq>0$)}, as predicted by Huygens' principle, the resulting wavefront (\new{cyan}) from the reflected waves \neww{at each point $\xm$} (\new{grey}) follows the specular direction. The simulation on the right takes into account wave propagation and interference, by solving \eref{eq:phasor_single_scattering} \neww{for all points $\xv$ in $\lV$} using Monte Carlo integration.
    } 
    \label{fig:hf_reflection}
    \Description[A planar diffuse surface reflects modulated light specularly]{In the real domain, light is scattered in all directions. In the computational domain, with only modulated components, an incoming planar wavefront is reflected specularly as predicted by Huygens' principle.}
\end{figure}

This is shown in \fref{fig:hf_reflection}a; we illustrate the capture process \new{(real domain)} of incoming light from $\xl$ \neww{that reaches points $\xm$ on a diffuse surface $\lM$} which, as the simulation on the right shows for points $\xv$ in the volume $\lV$, reflects light isotropically in all directions. \neww{In \fref{fig:hf_reflection}b, we illustrate the wave-based computational light transport of the same scene}, ignoring unmodulated light (frequency $\fq=0$) and considering only the part of the signal containing transient-modulated intensity (i.e., choosing a Fourier component with frequency $\Omega>0$). \neww{Also in \fref{fig:hf_reflection}b, the first schematic shows a planar light wavefront (red) at the time that it reaches the surface. The second schematic shows the reflected wavefront (cyan) \NEW{at a later time instant,} resulting from the superposition of spherical wavefronts (grey) at points $\xm$ in $\lM$. This} produces a specular reflection, as predicted by Huygens' principle, shown in the simulation on the right. Both simulations on the right of \fref{fig:hf_reflection} have been generated using Monte Carlo integration, using standard ray optics (\fref{fig:hf_reflection}a) and wave optics as described by \eref{eq:phasor_single_scattering} (\fref{fig:hf_reflection}b).

\begin{figure}[t]
    \centering
    \captionsetup{skip=3pt}
    \def\svgwidth{\columnwidth} 
    \begin{small}
    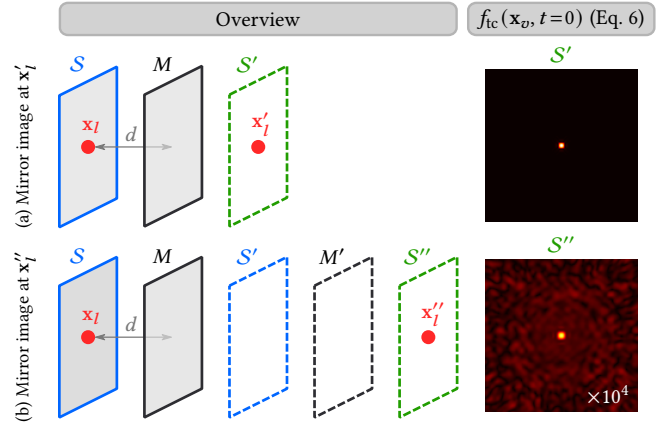
    \end{small}
    \caption{\neww{(a)} A planar surface $\lM$ coplanar to the aperture $\lS$ at a distance $d$ creates a mirror image \NEW{at} $\xlp$ of the \NEW{illuminated point} $\xl$ that we image from a third-bounce reconstruction method at a distance $2d$\new{; we show $\ftc(\xv, t=0)$ for points $\xv$ in the \NEW{plane $\lSp$, the} mirror image of $\lS$.} \neww{(b)} Since the \NEW{planar relay surface with} aperture $\lS$ also behaves as a mirror in the wave domain, another reflection of $\xl$ appears at a distance $4d$\new{, only changing the imaged volume to points $\xv$ in \NEW{the plane} $\lSpp$} (enhanced by a factor of $10^4$ for visualization purposes). Both images use $\wl_c = \sigma = \unit{3}{\centi\meter}$.}
    \label{fig:infinity_mirror}
    \Description[Two mirror surfaces create multiple mirror images of the illuminated point]{Top: Imaging at distance 2d shows a very bright spot in the middle; the rest of the image is black. Bottom: Imaging at distance 4d shows a bright spot in the middle which is clearly distinguishable, although the image is noisier.}
\end{figure}

\paragraph{\new{Infinity mirror experiment}}
\label{sec:recursivemirror}
To further illustrate how these specular reflections take place in a NLOS scenario, we set up a simple \new{simulated} scene made up of a diffuse hidden surface $\lM$ in front of the relay surface with aperture $\lS$\NEW{, at a distance $d$} (see \fref{fig:infinity_mirror}). \new{We illuminate a point $\xl$ on the relay surface using a laser device, and obtain the impulse response $H(\xl, \xs, t)$ at \NEW{points $\xs$ in} $\lS$ on the relay surface using transient rendering simulations~\cite{royo2022non, jarabo2014framework}.}
\neww{The relay surface (with aperture $\lS$) and $\lM$ are planar diffuse surfaces, thus behave like virtual mirrors in the computational wave domain. Looking at $\lM$, \NEW{
\NEWW{light emitted from $\xl$ is reflected by $\lM$, forming a mirror image of $\xl$ \emph{behind} $\lM$ like any conventional mirror}.
To capture in-focus images of the mirror reflection of $\xl$, we place the focal plane of the virtual camera \emph{behind} $\lM$.}
\NEWW{In particular,} we use the impulse response $H(\xl, \xs, t)$ and implement a transient camera model (\eref{eq:RSD_freq_tc}) to obtain $\ftc(\xv, t)$. We \neww{compute the \newww{time-resolved} image $\ftc(\xv, t)$} on the specific locations $\xv$ where the mirror images are formed. In \fref{fig:infinity_mirror}a\NEW{, we place the focal plane of the camera at a distance $2d$, on a plane $\lSp$ (green) which denotes}
the mirror image of $\lS$ \textit{behind} $\lM$. \NEW{The plane $\lSp$ also contains} the mirror image \NEW{at} $\xlp$ of $\xl$ produced by $\lM$\NEW{, captured in our result on the right as a bright spot.}}
\neww{
\newww{
As stated in \sref{sec:time-resolved-models}, each frame \NEW{at $t$ of $\ftc(\xv, t)$} combines points $\xv$ at different times \NEW{in the hidden scene}: the transient camera model captures events in the frame \NEW{at} $t$ equal to the time of flight from $\xl$ and the actual scene element corresponding to the mirror image at $\xlp$, which is also $\xl$ in this case. Consequently, the mirror image at $\xlp$ is captured in the frame at $t=0$.}}

We can image a higher-order mirror reflection pushing this effect even further, as shown in \neww{\fref{fig:infinity_mirror}b}. Since the diffuse relay surface that contains the aperture $\lS$ also behaves like a mirror in the computational wave domain, \neww{the specular interactions between $\lM$ and $\lS$ create additional mirror reflections of the \NEW{illuminated point} $\xl$ at locations further behind $\lM$}. 
\neww{This effect is analogous to the real situation where we observe multiple reflections of an object placed between two confronted, real mirrors. We showcase this effect in \fref{fig:infinity_mirror}b, using the same impulse response $H(\xl, \xs, t)$ and implementing the same transient camera model (\eref{eq:RSD_freq_tc}), but 
\newww{adjusting the focal \NEW{plane at a distance $4d$} to match}
points $\xv$ in the \NEW{plane} $\lSpp$. Looking at the computed image on the right, this yields a clear but dimmer spot that corresponds to the second mirror image at $\xlpp$ of the illuminated point $\xl$.}

\new{\paragraph{Computational wavelength}
The computational specular behavior of a real diffuse surface is explained by wave optics (illustrated in \fref{fig:hf_reflection}), and depends on the computational wavelengths $\lambda$ used through the imaging process ($\lambda$ is the inverse of the imaging frequencies $\fq$).
In practice, the computational wavelengths used through the imaging process depend on the frequency spectrum of the illumination function $\Pf(\xl,\fq)$. In our work, we determine the spectrum of frequencies $\fq$ of the imaging process through the central wavelength $\lambda_c$ and standard deviation $\sigma$ in \eref{eq:virtual_illumination}. 
The choice of these frequencies introduces a trade-off\neww{: lower values of $\lambda_c$ and $\sigma$ can properly image geometric features with more detail, but they may} also introduce unwanted high-frequency noise.
Following previous works, we use values for $\wl_c$ from $\unit{3}{\centi\meter}$ to $\unit{14}{\centi\meter}$, and values for $\sigma$ proportional to $\wl_c$. We specify the particular values used in each experiment. For reference, all experiments share the same aperture size of $\unit{2 \times 2}{\meter}$. Within this range of values, planar surfaces behave specularly during the NLOS imaging process, which we leverage to address different challenges of NLOS imaging methods.  We do not use wavelengths larger than this range, \NEW{as these may degrade the specular behavior of surfaces} due to the ratio between the surface size and the wavelength.}

\begin{figure}[t]
    \centering
    \captionsetup{skip=-1pt}
    \def\svgwidth{\columnwidth} 
    \begin{small}
    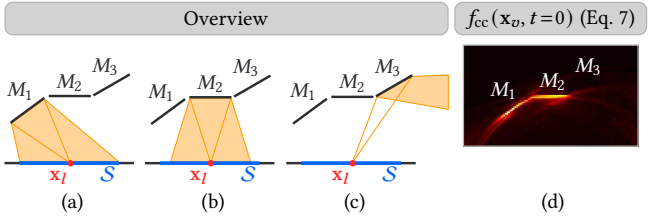
    \end{small}
    \caption{The visibility of surfaces in NLOS setups can be explained on the basis of their specular behavior in the wave domain. Our example follows the original missing cone explanation and uses the image (d) from work by \citet{Liu2019analysis}\neww{, obtained using the confocal camera model $\fcc$}. From an illuminated point $\xl$ in the relay surface, the visibility of the three surfaces $\lM_{1-3}$ depends on their position and orientation with respect to such relay surface (which also acts as \NEW{the} aperture $\lS$ \NEW{of a virtual camera}). As (a-c) show, only the reflected wavefronts from $\lM_1$ and $\lM_2$ reach the aperture $\lS$. As a result (d), the imaging system cannot see surface $\lM_3$.}
    \Description[Surface visibility explained using virtual mirrors]{(a-c) Show a top view of a scene with three surfaces positioned and oriented differently. The second surface is coplanar to the relay surface. The first and third surfaces are both oriented at a small angle, but only the first one is positioned such that it is facing the relay surface.}
    \label{fig:missing-cone}
\end{figure}

\new{\paragraph{Missing-cone problem through virtual mirrors} The frequency-space specular behavior of diffuse surfaces allows us to intuitively explain the missing-cone problem from \neww{our virtual mirrors} perspective:}
\neww{for a point $\xv$ in the hidden scene, if light from $\xl$ does not reach any point \NEW{$\xs$} in the aperture $\lS$ after a specular reflection in $\xv$, then the point $\xv$ is inside the null-reconstruction space of third-bounce imaging methods and cannot be reconstructed.}
\new{\fref{fig:missing-cone} \neww{illustrates this for third-bounce methods} \NEW{using} a scene with three diffuse surfaces $\lM_{1-3}$. However, only two \NEW{surfaces} are visible on the image shown in \fref{fig:missing-cone}d, computed using existing third-bounce NLOS imaging methods. Wave propagation in the computational NLOS imaging domain is illustrated in \fref{fig:missing-cone}a to \fref{fig:missing-cone}c for $\lM_1$ to $\lM_3$, respectively.} The reflected wavefronts from surfaces $\lM_1$ and $\lM_2$ reach the sensor $\lS$. However, given its particular position and orientation, this is not the case for $\lM_3$ and thus cannot be imaged. The surface $\lM_3$ is said to be \new{inside the null-reconstruction space}.

\new{Even if a surface is inside the null-reconstruction space \neww{of third-bounce imaging methods} (such as $\lM_3$ in \fref{fig:missing-cone}), we observe that the combination of several surfaces may produce higher-order illumination bounces that actually do reach $\lS$.}
\new{In \sref{sec:contribution-2:goal-2}, we show how to leverage such higher-order illumination to infer and to directly image objects that \new{are inside} the null-reconstruction space of third-bounce methods. We achieve this by analyzing mirror images of other objects produced by the surface inside the null-reconstruction space to infer its position and orientation. Beyond inference, we provide a procedure to translate our imaging system to a secondary surface that directly observes the target object.}

Moreover, in \sref{sec:contribution-2:goal-1} we show another way to leverage virtual mirrors in NLOS imaging, and image objects hidden behind \emph{two} corners using existing imaging models. For this we rely on fifth-bounce specular reflections in the computational domain to image the space behind \NEW{virtual mirror} surfaces, where the mirror image of such objects would appear.
\section{Addressing the missing cone}
\label{sec:contribution-2:goal-2}

\begin{figure*}[t]
  \centering
  \captionsetup{skip=6pt}
  \def\svgwidth{0.9\textwidth} 
  \begin{small}
  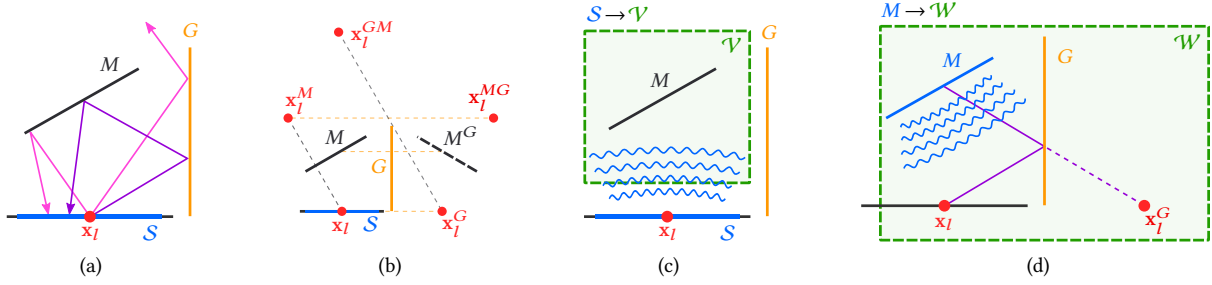
  \end{small}
  \caption{
  Illustration of our \newww{methodology that targets} \new{surfaces inside the null-reconstruction space of} third-bounce NLOS imaging methods by using higher-order illumination.
  (a) Example NLOS scene with an imaging aperture $\lS$, an illuminated point $\xl$, and surfaces $\lG$ and $\lM$. Surface $\lG$ is \new{inside the null-reconstruction space of} classic NLOS imaging methods due to lost third-bounce paths (pink). We use higher-order bounces (purple), which reach $\lS$ bouncing on both $\lM$ and $\lG$. \neww{Note that both pink and purple paths do not show all three and four bounces, respectively.}
  (b) Surfaces $\lM$ and $\lG$ produce mirror reflections of $\xl$ and $\lM$ after one or two specular reflections, denoted by their superscripts. We infer the position and orientation of $\lG$ as the plane between a point and its mirror image observed by the camera. For example, points $\xl$ and $\xlG$, or points $\xlM$ and $\xlMG$ are the reflection of each other \NEW{produced by} the surface $\lG$, denoted by the dotted lines in between.
  (c) \newww{\NEW{Our} direct imaging \NEW{procedure}:}
  \newww{To avoid inference ambiguities, we instead propose a procedure} to directly image $\xlG$ and $\lG$ by translating the aperture from $\lS$ to $\lM$. For this, we obtain the transient response at points $\xv \in \lV$, $\ftc(\xv, t)$, in a region $\lV$ (green) containing $\lM$, and evaluate $\ftc(\xv, t)$ at points $\xv \in \lM$.
  (d) We then implement imaging models at a secondary aperture $\lM$ (blue) that observes elements in \NEW{the volume} $\lW$ (green) that were \new{inside the null-reconstruction space} from $\lS$. Using $\lM$ as aperture, $\xlG$ \NEW{is captured by} the transient camera, and $\lG$ \NEW{is captured by} the confocal camera (paths $\xl \rarr \lG \rarr \lM$, purple).}
  \label{fig:goal2-overview}
  \Description[The reflection of the light source is used to estimate a virtual mirror surface]{A top view of our scene contains three planes: the relay surface, which contains the virtual aperture S; a plane G which is perpendicular to the relay surface; and a plane M which contains three- and four-bounce illumination. The procedure describes how to image the plane G using M as a secondary aperture.}
\end{figure*}

\begin{figure}
\captionsetup{skip=11pt}
\def\svgwidth{0.9\columnwidth} 
\begin{small}
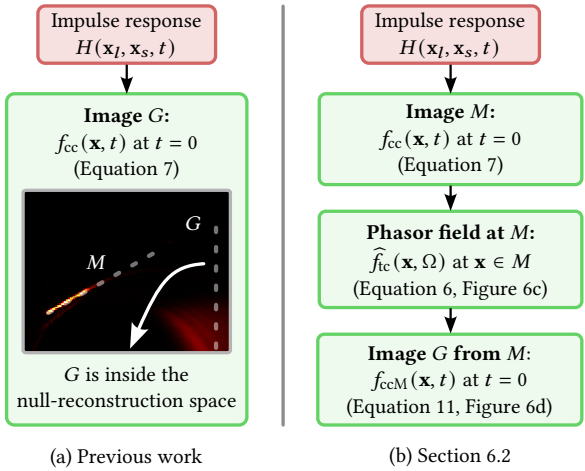
\end{small}
\caption{\new{Flowchart of the computational procedures applied to the impulse response $H(\xl, \xs, t)$ captured \NEW{for the scene} in \fref{fig:goal2-overview}a. (a) \newww{Following existing procedures}, trying to image $\lG$ using the confocal camera model $\fcc(\textbf{x}, t)$\NEW{, the frame at} $t=0$ shows that $\lG$ is inside the null-reconstruction space. (b) \newww{In \sref{sec:secondaryaperture} we propose a procedure to address this problem.} We use \NEW{the} confocal camera \NEW{model} with its aperture at $\lM$ instead of $\lS$, \NEW{denoted as $\fccM$,} which allows to image $\lG$ directly. For this second aperture at $\lM$, we compute the phasors \NEW{$\Pf(\textbf{x}, \fq) \equiv \fftc(\textbf{x}, \fq)$} for selected points $\textbf{x} \in \lM$ \NEW{using the transient camera model.}}}
\label{fig:goal2-diagram}
\Description[The image captures M but not G]{Using the confocal camera model with aperture S captures M but not G}
\end{figure}

\new{In the following, we propose \newww{a procedure} to address the missing-cone problem. Our key insight is that we can obtain information of surfaces inside the null-reconstruction space by analyzing the fourth-bounce illumination \NEW{that reaches} the relay surface through multiple interreflections with other surfaces in the hidden scene. First, we \newww{show how to analyze elements \NEW{captured} in time-resolved images of the hidden scene} to infer the position and orientation of surfaces inside the null-reconstruction space of third-bounce methods. \newww{As this inference is limited by ambiguities, we propose a procedure} to directly image such surfaces without requiring inference. All experiments in this section rely on simulated data using transient rendering~\cite{royo2022non, jarabo2014framework}.}

\paragraph{\new{Problem statement}} \fref{fig:goal2-overview}a shows an example scene, consisting of two surfaces $\lM$ and $\lG$, the NLOS imaging aperture $\lS$, and the illuminated point $\xl$. \neww{In the real domain, some light paths will bounce on the illuminated point $\xl$, then on $\lG$ and finally on a point in the aperture $\lS$, for a total of three bounces. However, }\new{in the NLOS computational domain, $\lG$ reflects third-bounce illumination specularly, away from $\lS$ (pink\neww{, note that not all three bounces are shown}). Therefore, imaging $\lG$ is limited if using third-bounce methods. \fref{fig:goal2-diagram}a illustrates the procedure of third-bounce methods, showing \NEW{that} $\lG$ is inside the null-reconstruction space.}

\new{Our key observation is that \NEW{four}-bounce paths (purple\neww{, note that not all four bounces are shown}) that reach $\lS$ through a specular-like reflection on $\lM$ ($\xl \rarr \lG \rarr \lM \rarr \lS$) provide valuable information about $\lG$. 
In particular, \newww{we show how to analyze illumination in time-resolved images of the scene} to infer the position and orientation of $\lG$ through mirror images of scene elements \NEW{produced} by $\lG$ (\sref{sec:mirrorimages}), and then propose a procedure to directly image $\lG$ by creating secondary apertures at surfaces that are visible by third-bounce methods such as $\lM$ (\sref{sec:secondaryaperture}).}

\subsection{Inferring surfaces from mirror images}
\label{sec:mirrorimages}
In this section, we show how to infer the position and orientation of $\lG$ \new{from the impulse response function $H(\xl, \xs, t)$}. The key insight is that if we can observe mirror images of scene elements created by $\lG$, it is because $\lG$ lies on the plane between such scene element and its mirror image, just like what happens with a real mirror. \fref{fig:goal2-overview}b points the location of mirror images of the illuminated point $\xl$ and $\lM$, \NEW{produced} by $\lM$ and $\lG$ in our example scene. Superscripts \NEW{denote the surface (or surfaces, in order) that produces the mirror images of each scene element:}
$\xl$ is mirrored by $\lM$ and $\lG$ at $\xlM$ and $\xlG$, respectively; 
$\lM$ is mirrored by $\lG$ at $\lMG$;
$\xlG$ is mirrored by $\lM$ at $\xlGM$;
\NEW{and} $\xlM$ is mirrored by $\lG$ at $\xlMG$.

\begin{figure}[t]
  \centering
  \captionsetup{skip=0pt}
  \begin{small}
  \def\svgwidth{\columnwidth}
  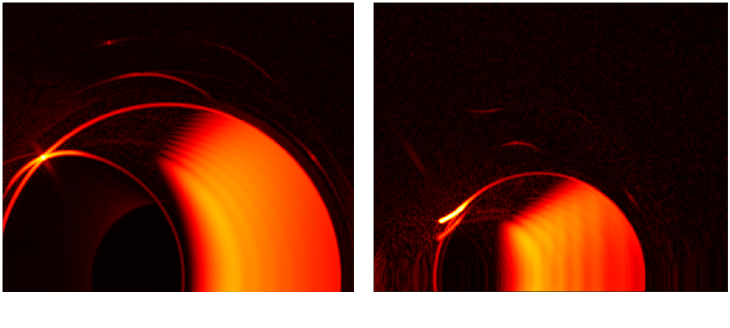
  \end{small}
  \caption{(a) Mirror images \NEW{at} $\xlM$, $\xlGM$ and $\xlMG$ of $\xl$ produced by surfaces $\lM$ and $\lG$, computed with the transient camera $\ftc(\textbf{x}, t)$ (\eref{eq:RSD_freq_tc}) \NEW{for points $\textbf{x}$ that cover the region depicted in \fref{fig:goal2-overview}b}. \newww{Although these reflections happen at different times, our computed time-resolved image composes them in the same frame at $t=0$.} 
  (b) When light \NEW{from $\xl$} reaches the center point $\overline{\mathbf{x}}_m$ of $\lM$ at $\overline{t}_m=\norm{\overline{\mathbf{x}}_m - \xl}/c$, we can see both $\lM$ and its mirror image $\lMG$ produced by $\lG$. Both images are computed using $\lambda_c = \sigma = \unit{3}{\centi\meter}$.} 
  \label{fig:mirror_images_from_S}
  \Description[The camera captures three reflected bright spots from xl and one reflected surface from M]{The transient camera captures three reflections of the light source at xl; also the surface M and its reflection on G. }
\end{figure}

\paragraph{Observing mirror images from $\lS$} 
\neww{Previously, we described the location of mirror images of the illuminated point $\xl$ produced by surfaces in the hidden scene, as shown in \fref{fig:goal2-overview}b.}
\neww{To show which of these mirror images are visible from the aperture $\lS$ and use them to infer the location and orientation of $\lG$}, we implement a transient camera model $\ftc$ (\eref{eq:RSD_freq_tc}), and evaluate it at different time instants.
\new{In the resulting time-resolved image $\ftc(\xv,t)$, light emitted from the illuminated point $\xl$ \newww{is captured in the frame} at $t=0$. \neww{We show \NEW{the frame} $\ftc(\xv,t=0)$ in \fref{fig:mirror_images_from_S}a, \NEW{which captures} several bright spots at the locations of the mirror images of $\xl$ that are visible from aperture $\lS$ \NEW{over a volume that covers the whole scene.} In particular, the}}
 aperture $\lS$ can observe the mirror images \neww{at} $\xlM$ (produced by \NEW{three}-bounce paths $\xl\rarr\lM\rarr\lS$), \neww{at} $\xlMG$ and \new{at} $\xlGM$ (both produced by \NEW{four}-bounce paths $\xl\rarr\lM\rarr\lG\rarr\lS$ and $\xl\rarr\lG\rarr\lM\rarr\lS$).
We cannot, however, observe the mirror image \neww{at} $\xlG$, since \NEW{three}-bounce paths from $\xl$ to $\lG$ do not reach $\lS$ in the computational domain (\fref{fig:goal2-overview}a, pink).
\neww{Evaluating $\ftc(\xv,t)$ at \NEW{frames with} $t>0$ may show other scene elements and their mirror images, similarly to $\xl$, based on the time of flight from $\xl$ to \NEW{each scene element.} 
For example,
\new{light from the \NEW{illuminated point} $\xl$ will reach the central point of $\lM$, denoted as $\overline{\mathbf{x}}_m$ (\fref{fig:mirror_images_from_S}a), at the time of flight $\overline{t}_m = \norm{\overline{\mathbf{x}}_m - \xl}/c$.} We can therefore identify reflections at points near the center of the plane $M$ by \NEW{looking at the frame at $t=\overline{t}_m$ of} $\ftc(\xv, t)$ (\fref{fig:mirror_images_from_S}b), \NEW{which captures}}
not only $\lM$ and $\lMG$, but many other mirror images produced by $\lM$ and $\lG$.

\neww{Both images in \fref{fig:mirror_images_from_S} show bright areas outside the points or surfaces mentioned before. This is mainly \NEW{because} the impulse response function $H(\xl, \xs, t)$ \NEW{combines coupled information from paths of different bounces and optical lengths}, which introduces out-of-focus, low-frequency artifacts at each imaged location $\xv$.}
\neww{Also, note that during the NLOS imaging process we only know the location of the illuminated point at $\xl$, \NEW{and} we do not have any prior knowledge of other \NEW{hidden} scene elements. Consequently, there exist ambiguities when identifying the \NEW{captured} bright spots as mirror images of $\xl$,} so we cannot guarantee that e.g., $\xlMG$ is a mirror image of $\xlM$; instead e.g., there may be two physical surfaces located at $\lMG$ and $\lM$, respectively. 

\new{\paragraph{Inferring the position and orientation of $\lG$} By assuming that $\xlMG$ is a mirror image of $\xlM$
---i.e., $\xl$ has undergone two reflections \NEW{produced} by $\lM$ and \NEW{then} $\lG$---the surface $\lG$ that produced $\xlMG$ should lie in the perpendicular plane between $\xlM$ and $\xlMG$.} We obtain a point $\xg$ on such plane and its normal vector $\normalg$ as
\begin{equation}
\xg = \frac{\xlM+\xlMG}{2}, \quad \normalg = \frac{\xlM - \xlMG}{\norm{\xlM - \xlMG}},
\label{eq:infer-cg-ng}
\end{equation}
\new{which define the position and orientation of $\lG$, respectively.}
We could also infer \new{the position and orientation} of $\lG$ from $\lM$ and its mirror reflection $\lMG$. 
Such inferences require assumptions on the number of reflections undergone by the observed patterns (\fref{fig:mirror_images_from_S}). For this particular inference, $\xlM$ can be identified as the reflection of $\xl$ from the visible orientation of $\lM$, and \NEW{$\xlMG$} cannot be produced by $\lM$, so we assume it is a second mirror reflection by another hidden surface. 

In general, there are ambiguities on recognizing the source of every identified reflection, which may introduce errors when inferring hidden surfaces. 
\newww{To avoid errors due to such ambiguities}, in the following we propose a procedure to directly image $\lG$ without requiring inference.

\subsection{Imaging surfaces from secondary apertures}
\label{sec:secondaryaperture}
\new{Here we show a procedure to directly image $\lG$ using fourth-bounce illumination. The key idea is that, while the surface $\lG$ is inside the null-reconstruction space of imaging systems created at $\lS$ (\fref{fig:goal2-overview}a, pink), $\lG$ is not inside the null-reconstruction space of imaging systems created at $\lM$, since there exist three-bounce paths $\xl \rarr \lG \rarr \lM$ that reach $\lM$ (\fref{fig:goal2-overview}d, purple). Based on this observation, we show how to computationally translate our imaging system from $\lS$ to $\lM$ (\fref{fig:goal2-overview}c) to directly image $\lG$ using $\lM$ as a secondary aperture (\fref{fig:goal2-overview}d).}

\new{Our procedure is illustrated in \fref{fig:goal2-diagram}b: first, how to obtain a phasor field at $\lM$, and then how to use this response to generate computational cameras with the aperture located at $\lM$. The rest of this section describes this procedure in detail along with its results.}

\paragraph{\new{Phasor field at $\lM$}} \new{The phasor-field formulation uses phasors $\Pf(\xs, \fq)$ at points $\xs$, which encode the response of the scene to the illumination function, to implement imaging models with a camera aperture $\lS$. To translate the camera aperture from points $\xs$ to points $\xm \in \lM$, we need to compute their corresponding phasors $\Pf(\xm,\fq)$.
The transient camera model $\ftc$ achieves this goal, since $\Pf(\xm, \fq) \equiv \fftc(\xm, \fq)$ propagates phasors from all points $\xs$ to \NEW{each point} $\xm$. To determine where $\lM$ is, we first implement a confocal camera model \neww{to capture} $\fcc(\xv,t)$ from $\lS$---equivalent to existing NLOS imaging models---\neww{to image all points $\xv$ in the region $\lV$ (\fref{fig:goal2-overview}c). The computed \NEW{frame at $t=0$} can be seen in \fref{fig:goal2-diagram}a. We can \NEW{estimate} all points $\xm$ in $\lM$ by thresholding the image $\fcc(\xv, t=0)$.
} We then implement a transient camera model \neww{to obtain} $\fftc(\xm,\fq)$, yielding a phasor field at $\lM$ that can be used to implement new imaging systems at $\lM$.}

\new{\paragraph{$\lM$ as a secondary aperture}
Based on RSD propagation principles, we implement a lens at $\lM$ which focuses the phasors $\fftc(\xm,\fq)$ at points $\xw$ in a volume $\lW$ that contains $\lG$ (\fref{fig:goal2-overview}d), equivalent to what the transient camera model does. In practice, we include an RSD propagator from points $\xm$ to points $\xw$, yielding the transient camera model with aperture $\lM$ (denoted as $\ftcM$) as
\begin{align}
    \fftcM(\xw, \fq) = \int\limits_{M} \frac{e^{i k \norm{\xw-\xm}}}{\norm{\xw-\xm}}\fftc\left(\xm, \fq\right) \diff \xm.
    \label{eq:second_RSD}
\end{align}}
The \new{time-domain version $\ftcM(\xw, t) = \mathcal{F}^{-1}\left\{ \fftcM(\xw, \fq) \right\}$} represents a time-resolved image of the scene as captured from $\lM$.
\NEW{The frame at $t=0$ of} $\ftcM(\xw, t)$ \NEW{captures} the initial light up of the \NEW{illuminated point} $\xl$ and its mirror image at $\xlG$.
Similarly to the mirror images captured from $\lS$ (\fref{fig:mirror_images_from_S}), given $\xl$ and its mirror image $\xlG$ (\NEW{captured by the aperture at} $\lM$) we can infer the position and orientation of the surface $\lG$ that \NEW{produced} $\xlG$ (\eref{eq:infer-cg-ng}).

\new{\paragraph{Imaging $\lG$ from $\lM$} 
Instead of approximating $\lG$ through geometric inference, we \emph{directly} image $\lG$ by extending \eref{eq:second_RSD} to implement a confocal camera model at $\lM$. Under a single \NEW{illuminated point} $\xl$, this is equivalent to incorporating an RSD operator to \eref{eq:second_RSD} that propagates the phasor \NEW{$\Pf(\xm, \fq) \equiv \fftc(\xm, \fq)$} accounting for the distance between $\xl$ and $\xw$, yielding a confocal camera model with aperture $\lM$ (denoted as $\fccM$) as
\begin{align}
    \ffccM(\xw,\fq) & = \fftcM(\xw, \fq) \frac{e^{i k \norm{\xw-\xl}}}{\norm{\xw-\xl}} \nonumber \\
                    & = \int\limits_{\lM} \frac{e^{i k \left(\norm{\xw-\xm} + \norm{\xw-\xl}\right)}}{\norm{\xw-\xm}\norm{\xw-\xl}} \fftc(\xm, \fq) \diff \xm.
\label{eq:second_RSD_confocal}
\end{align}
}
\new{The time-resolved image $\fccM(\xw, t) = \mathcal{F}^{-1}\left\{ \ffccM(\xw, \fq) \right\}$ of this confocal camera model \NEW{captures} $\lG$ \NEW{in the frame at} $t=0$.} This follows the basis of classic NLOS reconstruction methods, which directly image $\lG$ from the time of flight of \NEW{three}-bounce paths $\xl \rarr \lG \rarr \lM$, but instead using \NEW{four}-bounce paths $\xl \rarr \lG \rarr \lM \rarr \lS$ in the captured impulse response $H(\xl, \xs, t)$.

\begin{figure}[t]
  \centering
  \captionsetup{skip=-3pt}
  \def\svgwidth{\columnwidth} 
  \begin{small}
  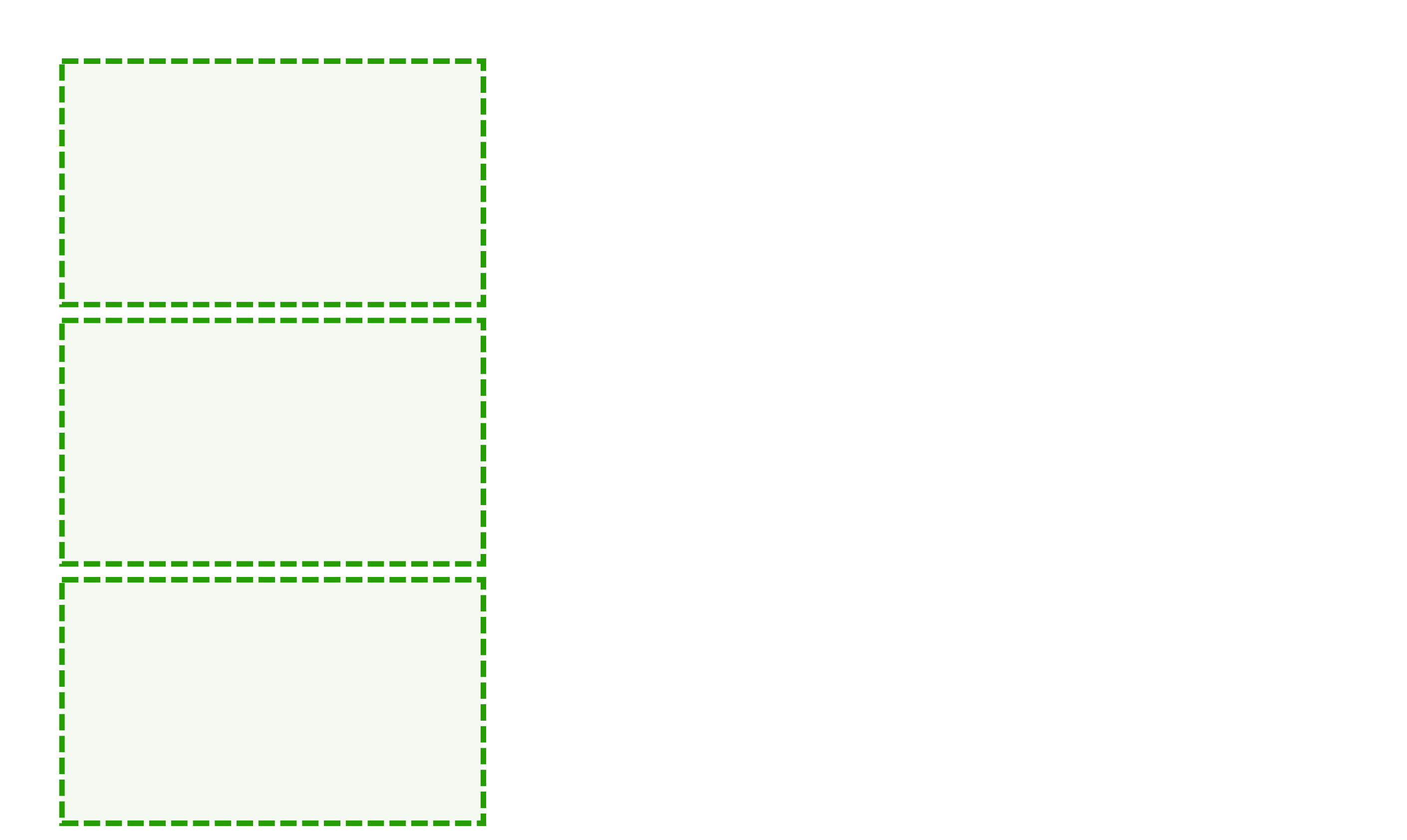
  \end{small}
  \caption{\textbf{Left column:} We rotate $\lG$ in \fref{fig:goal2-overview}a to show that it acts as a virtual mirror surface. In (a-c), $\lG$ is rotated $\unit{90}{\degree}$, $\unit{100}{\degree}$ and $\unit{80}{\degree}$ with respect to the relay surface. Rotating $\lG$ changes the position of the mirror image \NEW{at} $\xlG$ of the illuminated point $\xl$. We image all points $\xw$ in the volume $\lW$. \textbf{Middle column:} From the known position of $\xl$ and the computed position of $\xlG$, we can use the transient camera model with aperture $\lM$ \NEW{($\ftcM$)} and \NEW{infer the position $\xg$ and orientation $\normalg$ of surface $\lG$.}
  \textbf{Right column:} We directly image $\lG$ using the confocal camera model with aperture $\lM$ \NEW{($\fccM$)}. All images use $\wl_c = \sigma = \unit{3}{\centi\meter}$.}
  \label{fig:goal2-results}
  \Description{TODO.}
\end{figure}

\new{\subsection{Results}
We illustrate results of our \newww{procedure} using the scene in \fref{fig:goal2-overview}a for both inference based on $\xl$ and its mirror image $\xlG$, and direct imaging of $\lG$ in \fref{fig:goal2-results}. We use different orientations of the plane $\lG$ to show how it affects the resulting mirror image $\xlG$ and the direct image of $\lG$.
In the left column we show an overview of the scenes.
Note that while inference (\sref{sec:mirrorimages}) and direct imaging (\sref{sec:secondaryaperture}) are two separate procedures, here we illustrate inference using the mirror image \NEW{at} $\xlG$ \new{captured} through the secondary aperture at $\lM$ of the direct imaging procedure (\sref{sec:secondaryaperture}).}

\new{We show results for inference, identifying both $\xl$ and $\xlG$ in the computed images $\ftcM(\xw, t)$ (\eref{eq:second_RSD}) at $\xw \in \lW$ and $t=0$ (\fref{fig:goal2-results}, middle column). From $\xl$ and $\xlG$ we infer a point $\xg$ and the normal $\normalg$ of a plane corresponding to the surface $\lG$.}

\new{To illustrate our direct imaging procedure, we compute \eref{eq:second_RSD_confocal} to directly image $\lG$, evaluating $\fccM(\xw, t)$ at $\xw \in \lW$ and $t=0$ (\fref{fig:goal2-results}, right column).
This yields a clear image \neww{of the surface $\lG$ which} is entirely on the null-reconstruction space of third-bounce methods (\fref{fig:goal2-diagram}a). Note only part of $\lG$ is visible, the rest is inside the null-reconstruction space of the imaging system at $\lM$ as some fourth-bounce illumination paths do not reach $\lS$ through specular bounces on $\lM$ in the computational domain. }
\neww{The results also show a bright region near $\xl$. In this case, we use our fourth-bounce imaging method, but the impulse
response function $H(\xl, \xs, t)$ \newww{combines coupled information from paths of different bounces and optical lengths}, which translates into out-of-focus, low-frequency artifacts.}

\section{Looking around two corners}
\label{sec:contribution-2:goal-1}
Here we leverage our virtual mirror reflections in the wave domain to image objects around \emph{two} corners.
\neww{The fundamental idea of our approach is exploiting our observation that diffuse planar surfaces behave like virtual mirrors (\sref{sec:light_interference}); we image the region where their mirror image is formed using only the confocal camera model (\eref{eq:RSD_freq_cc}). Note this imaging model is similar to those used by third-bounce NLOS imaging methods\newww{, analogous to the procedure detailed in \fref{fig:goal2-diagram}a}. However, unlike \newww{previous work}, we exploit fifth-bounce illumination in the impulse response function $H(\xl, \xs, t)$ by shifting the imaged region based on our observed behavior of virtual mirrors. 
In the following we detail this procedure using simulated data.}

\begin{figure}[t]
  \centering
  \captionsetup{skip=0pt}
  \def\svgwidth{\columnwidth} 
  \begin{small}
  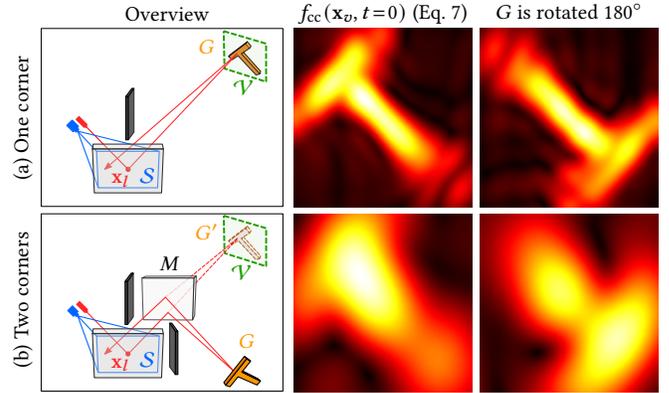
  \end{small}
  \caption{(a) A T-shaped object is hidden behind one corner. Focusing the confocal camera at points $\xv$ in $\lV$ (using $\wl_c = \unit{5}{\centi\meter}, \sigma = \unit{6}{\centi\meter}$) uses three-bounce illumination produced by $\lG$, and the T-shaped object is visible in the image when $t=0$. (b) Our two-corner imaging setup. None of the points $\xs$ in $\lS$, or $\xl$, have a direct line of sight towards $\lG$, that is, three-bounce illumination cannot contain information about $\lG$. Focusing the confocal camera at \NEW{points $\xv$ in} $\lV$ (using $\wl_c = \unit{12}{\centi\meter}, \sigma = \unit{14}{\centi\meter}$) \NEW{captures} the mirror image \NEW{at} $\lGp$ of the geometry $\lG$ produced by the diffuse surface $\lM$. We also rotate the T-shaped object by $\unit{180}{\degree}$, to show that it also affects the resulting image.}
  \label{fig:goal1-results}
  \Description[Imaging a T-shaped object around two corners]{Example scene where a diffuse surface acts like a virtual mirror that allows imaging a T-shaped object hidden behind two corners.}
\end{figure}

\new{We illustrate this with the simulated scene depicted in \fref{fig:goal1-results}b, composed by the relay surface with an illuminated point $\xl$ and the aperture $\lS$, a diffuse T-shaped object hidden behind two corners \NEW{denoted as} $\lG$ (which we aim to image), and a diffuse surface $\lM$ hidden behind a single corner, which behaves as a virtual mirror during the computational imaging process. We also place two black occluders to ensure that $\lM$ is not \NEW{directly} visible to the NLOS imaging device, and $\lG$ is not \NEW{directly} visible neither to the NLOS device nor to the imaging aperture $\lS$ \NEW{or illuminated point $\xl$}.}

\new{Our goal} is to image \new{the} object $\lG$.
\new{Due to the location and orientation of $\lM$ in our scene, the aperture $\lS$ and the object $\lG$ are at each other's specular direction with respect to $\lM$, \new{so that $\lM$ forms a mirror image of $\lG$ in the space behind $\lM$ \neww{that is captured from $\lS$}}. Specifically, we place the diffuse surface $\lM$ coplanar to the aperture $\lS$ with a lateral shift so it reflects light specularly towards $\lG$ and back to $\lS$ through another bounce in $\lM$. This creates fifth-bounce paths with the form $\xl \rarr \lM \rarr \lG \rarr \lM \rarr \lS$ (marked in red in the schematic of \fref{fig:goal1-results}b), which we leverage to image $\lG$. \newww{Also, note that 
in our setup $\lM$ is in the null-reconstruction space of third-bounce methods, as third-bounce specular paths do not reach $\lS$ in the computational domain. However, fifth-bounce specular paths actually reach $\lS$, and therefore we can image $\lG$ even when $\lM$ is in the null-reconstruction space.}}

\new{\neww{In our experiment setup, we first} obtain a simulated impulse response function $H(\xl,\xs,t)$ of the scene using transient rendering~\cite{royo2022non, jarabo2014framework} to mimic a real acquisition process of the relay surface.}
We then \new{implement a confocal camera model (\eref{eq:RSD_freq_cc}) and compute \NEW{the frame at $t=0$ of} $\fcc(\xv, t)$ \neww{at points $\xv$ in the imaged \NEW{plane on} $\lV$} (\fref{fig:goal1-results}b, green), where the mirror image} of $\lG$ \NEW{produced by} $\lM$ would be formed ($\lGp$ in the schematic). 
The computed images on \fref{fig:goal1-results}b show the result of this imaging process for two orientations of the T-shaped geometry, showing that the shape's structure is preserved on both, even after this second corner. For reference, we configure a single-corner scene (\fref{fig:goal1-results}a) by removing the surface $\lM$ and placing the object $\lG$ at the position \NEW{marked by} $\lGp$, where the mirror image should be formed for the two-corner case.
\neww{The images of the T-shaped object appear blurrier when imaged around two corners. Following our observations in \sref{sec:light_interference}, this effect is mainly caused by two factors. First, even if $\lM$ is perfectly diffuse, the mirror-like behavior of $\lM$ in the computational domain is not perfectly specular\newww{, and the resolution of the mirror images that $\lM$ produces is limited by diffraction.} Second, we use larger wavelengths on the two-corner case (i.e., values for $\lambda_c$ and $\sigma$ are higher in \fref{fig:goal1-results}b than \fref{fig:goal1-results}a). We discuss this mirror behavior and its effects further on \sref{sec:discussion}.}

\section{Results in real scenes}
\label{sec:real-captures}
In the following we illustrate and validate our methods in real  scenarios, imaging diffuse planar surfaces \new{inside the null-reconstruction space of} third-bounce methods in single-corner configurations, and then imaging scenes hidden behind two corners.

\paragraph{Hardware details}
\begin{figure}[t]
  \centering
  \captionsetup{skip=0pt}
  \def\svgwidth{\columnwidth} 
  \begin{small}
\begingroup%
  \makeatletter%
  \providecommand\color[2][]{%
    \errmessage{(Inkscape) Color is used for the text in Inkscape, but the package 'color.sty' is not loaded}%
    \renewcommand\color[2][]{}%
  }%
  \providecommand\transparent[1]{%
    \errmessage{(Inkscape) Transparency is used (non-zero) for the text in Inkscape, but the package 'transparent.sty' is not loaded}%
    \renewcommand\transparent[1]{}%
  }%
  \providecommand\rotatebox[2]{#2}%
  \newcommand*\fsize{\dimexpr\f@size pt\relax}%
  \newcommand*\lineheight[1]{\fontsize{\fsize}{#1\fsize}\selectfont}%
  \ifx\svgwidth\undefined%
    \setlength{\unitlength}{525.08567029bp}%
    \ifx\svgscale\undefined%
      \relax%
    \else%
      \setlength{\unitlength}{\unitlength * \real{\svgscale}}%
    \fi%
  \else%
    \setlength{\unitlength}{\svgwidth}%
  \fi%
  \global\let\svgwidth\undefined%
  \global\let\svgscale\undefined%
  \makeatother%
  \begin{picture}(1,0.50176494)%
    \lineheight{1}%
    \setlength\tabcolsep{0pt}%
    \put(0,0){\includegraphics[width=\unitlength,page=1]{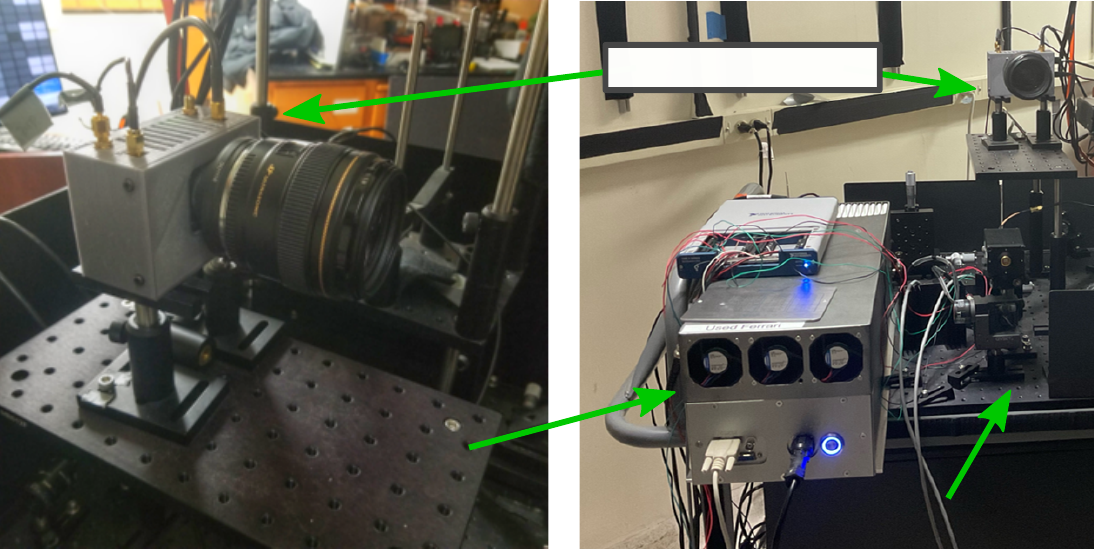}}%
    \put(0.67802395,0.43221344){\color[rgb]{0,0,0}\makebox(0,0)[t]{\lineheight{1.25}\smash{\begin{tabular}[t]{c}(a) SPAD array\end{tabular}}}}%
    \put(0,0){\includegraphics[width=\unitlength,page=2]{hardware-v2.pdf}}%
    \put(0.36892573,0.08728243){\color[rgb]{0,0,0}\makebox(0,0)[t]{\lineheight{1.25}\smash{\begin{tabular}[t]{c}(b) Laser\end{tabular}}}}%
    \put(0,0){\includegraphics[width=\unitlength,page=3]{hardware-v2.pdf}}%
    \put(0.84792371,0.02470917){\color[rgb]{0,0,0}\makebox(0,0)[t]{\lineheight{1.25}\smash{\begin{tabular}[t]{c}(c) Galvanometer\end{tabular}}}}%
  \end{picture}%
\endgroup%

  \end{small}
  \caption{Our hardware setup used to validate our \NEW{procedures} in real scenarios. (a) 16x16 Single Photon Avalanche Diode (SPAD) array \new{focused at \NEW{a point on} the relay surface.} (b) Laser source which can emit 35 picosecond pulses. (c) Two-mirror galvanometer \NEW{that guides the laser} to scan the relay surface.}
  \label{fig:hardware-setup}
  \Description[Hardware setup]{A picture of a SPAD array, a laser source and a galvanometer.}
\end{figure}
Our NLOS imaging system consists of a SPAD array sensor, a laser emitter and a two-mirror galvanometer (\fref{fig:hardware-setup}).
\new{The galvanometer guides the laser towards multiple points on the relay surface, while the detector is aimed at a fixed position on the relay surface.} 
A $\text{PM-1.03-25}^\text{TM}$ laser from Polar Laser Laboratories is used as an illumination source.
The laser is combined with a frequency doubler to emit $\unit{515}{\nano\metre}$ pulses with a maximum pulse width of $\unit{35}{\pico\second}$, an average power of $\unit{375}{\milli\watt}$, and at an average repetition rate of $\unit{5}{\mega\hertz}$. A two-mirror Thorlabs galvanometer (Thorlabs GVS012) is used to scan a relay surface at $\unit{1}{\centi\metre}$ spacing, with a total scan area \neww{around $\unit{1.9 \times 1.9}{\meter}$}.
Our detector is a 16x16 Single Photon Avalanche Diode (SPAD) array \cite{riccardoFastGated16162022} focused at a $\unit{7.1}{\centi\metre}$ by $\unit{4.7}{\centi\metre}$ area on the relay surface using a Canon EF $\unit{85}{\milli\metre}$ f/1.8 USM Lens. The temporal resolution of the array has a Full-Width at Half Maximum (FWHM) of around $\unit{60}{\pico\second}$ and a deadtime of less than $\unit{100}{\nano\second}$.
\new{All scene surfaces are diffuse expanded polystyrene foam and unfinished drywall, with no retroreflective properties. }

\begin{figure*}[t]
  \centering
  \captionsetup{skip=-2pt}
  \def\svgwidth{\textwidth} 
  \begin{small}
  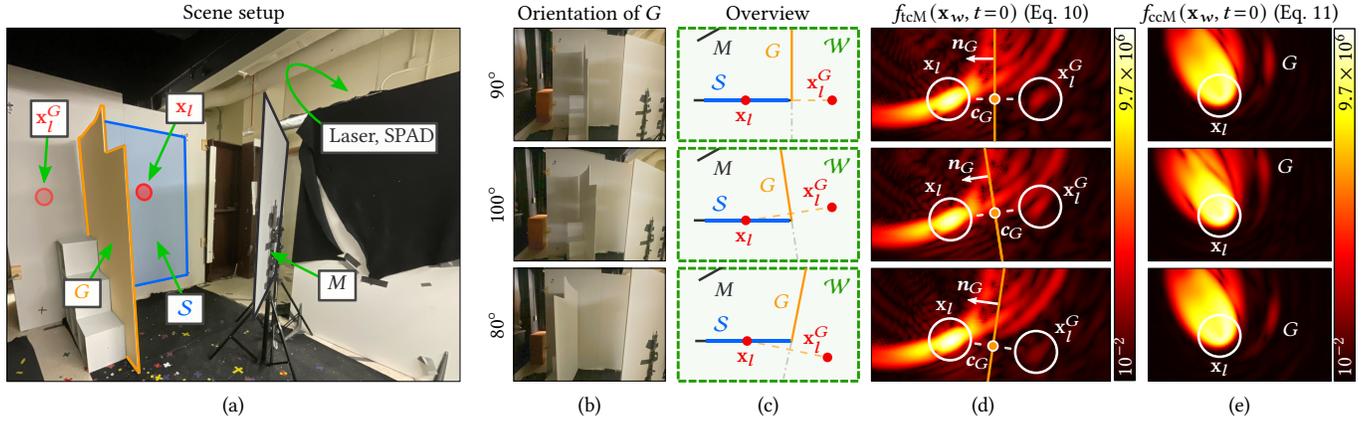
  \end{small}
  \caption{\new{Results for the inference of the position and orientation of surfaces inside the null-reconstruction space, and direct imaging of such surfaces (\sref{sec:contribution-2:goal-2}, same setup) using real captures.} (a) Photograph of the general scene setup. Laser and SPAD array are on the right. (b) Photographs of different orientations of the surface $\lG$ inside the null-reconstruction space, to be estimated for three independent experiments, one per row. (c) Overview of the scene setups (changing the orientation of $\lG$) and the imaged volume $\lW$ with points $\xw \in \lW$. (d) Inference of the \NEW{position $\xg$ and orientation $\normalg$ of surface} $\lG$, from the illuminated point $\xl$ and its \NEW{mirror image at} $\xlG$ \NEW{produced} by the virtual mirror surface $\lG$. Computed using $\wl_c = \sigma = \unit{7}{\centi\metre}$. (e) Imaging of the surface $\lG$ from the secondary aperture $\lM$, using $\wl_c = \unit{10}{\centi\metre}$ and $\sigma = \unit{7}{\centi\metre}$. Both the inferred \new{position and orientation of} $\lG$ and the direct image of $\lG$ taken from $\lM$ are a close match with respect to the capture setup in all three cases, even though it is not visible from the relay surface for classic NLOS imaging methods. \NEW{The colorbar is displayed in a logarithmic scale.}}
  \label{fig:goal2-real}
  \Description[Our estimation agrees with the rotation of the wall]{We estimate the position and orientation of three null-visibility surfaces, and image part of them}
\end{figure*}

\paragraph{\new{Addressing the missing cone}} We design a \new{scene similar to \fref{fig:goal2-overview}a to test our \newww{procedure to} address the \NEW{missing-cone problem} with our hardware setup. A photograph of the scene is displayed in  \fref{fig:goal2-real}a, with two \NEW{hidden} surfaces $\lM$ and $\lG$. Surface $\lG$ is in the null-reconstruction space of third-bounce methods since third-bounce illumination from $\lG$ falls outside our imaging aperture $\lS$ in the computational domain.}
\new{We aim to first infer the position and orientation of $\lG$, 
then to directly image $\lG$. We capture the impulse response function $H(\xl, \xs, t)$ for an illuminated point $\xl$ and points $\xs$ of the aperture $\lS$ at the relay surface. The surface $\lM$} produces fourth-bounce illumination at $\xs$ from our target diffuse surface $\lG$. We experiment with three different orientations of $\lG$ ($\unit{90}{\degree}$, $\unit{100}{\degree}$ and $\unit{80}{\degree}$) with respect to the relay surface, which is in all cases located at $\unit{50}{\centi\metre}$ from the illuminated point $\xl$ (in all three cases, $\lG$ cannot be imaged using existing NLOS algorithms). The surface $\lM$ is tilted at $\unit{30}{\degree}$ and separated $\unit{1.5}{\metre}$ from $\xl$ \NEW{at its center point}. \fref{fig:goal2-real}b and c show \NEW{photographs} and top-view schematics of the different orientations, respectively. 
\fref{fig:goal2-real}d shows how, for all three orientations, \neww{the position $\xg$ and orientation $\normalg$ of $\lG$} are accurately inferred from the illuminated point $\xl$ and its reflection $\xlG$ \new{\NEW{captured} with the transient camera from the aperture $\lM$ (\eref{eq:second_RSD}), in $\ftcM(\xw, t=0)$.}
For the $\unit{90}{\degree}$, $\unit{100}{\degree}$ and $\unit{80}{\degree}$ cases, our inferred $\xg$ is $\unit{6}{\centi\metre}$, $\unit{4}{\centi\metre}$ and $\unit{5}{\centi\metre}$ away from $\lG$, and $\normalg$ has an orientation error of $\unit{0.3}{\degree}$, $\unit{2.8}{\degree}$ and $\unit{0.1}{\degree}$, respectively. 
\neww{Additionally, to directly image plane $\lG$ we turn $\lM$ into a secondary aperture where we implement a} confocal camera  (\eref{eq:second_RSD_confocal}) to obtain the image $\fccM(\xw, t=0)$.
The results are shown in \fref{fig:goal2-real}e. \neww{Similar to \fref{fig:goal2-results}, the $\fccM$ model produces a bright region near $\xl$ due to \NEW{coupled} illumination in the impulse response $H(\xl, \xs, t)$.}

\begin{figure*}[t]
  \centering
  \captionsetup{skip=-2pt}
  \def\svgwidth{\textwidth} 
  \begin{small}
  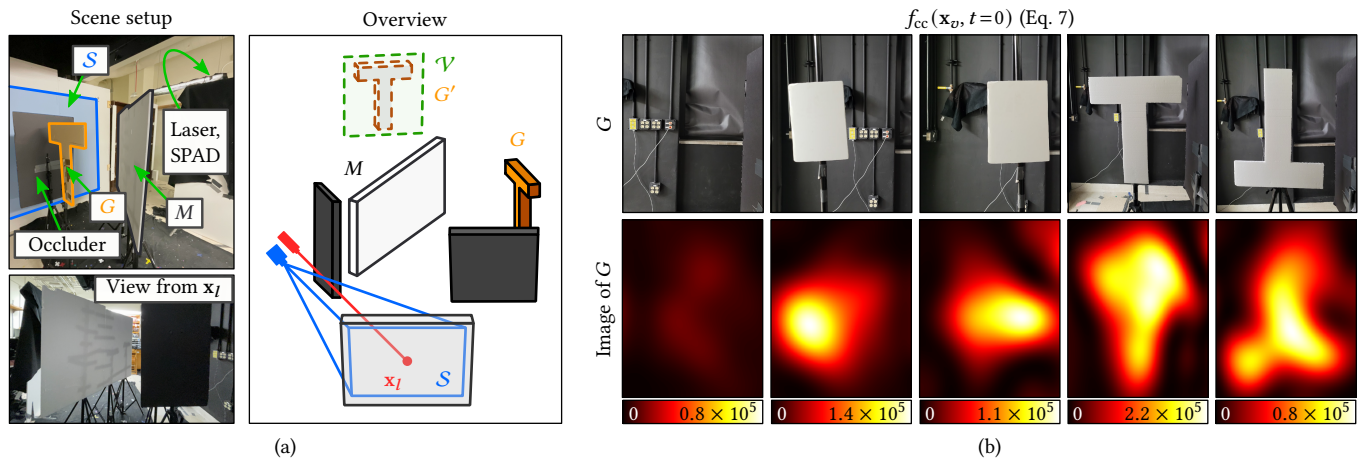
  \end{small}
  \caption{Results for two-corner imaging (\sref{sec:contribution-2:goal-1}, similar setup) from real captured data. (a) Photographs and overview of the setup. The geometry $\lG$ is hidden behind two corners: the relay surface with aperture $\lS$, and a diffuse surface $\lM$ that \NEW{is a virtual mirror in the computational domain}. \new{This diffuse surface $\lM$ is oriented so that the specular reflection from $\xl$ reaches $\lG$, and the specular reflection from $\lG$ reaches $\lS$. The geometry $\lG$ is not directly visible from $\xl$ or any point in the aperture $\lS$ as it is covered by an occluder. (b) \textbf{Top:} Photographs of the objects hidden around two corners.} \textbf{Bottom:} Imaging results for different geometries $\lG$, one per column, \NEW{placing the focal plane of the virtual camera at $\lV$.} We can clearly identify shape, locations and orientations despite the geometries being hidden after a second corner. Computed using $\wl_c = \unit{14}\centi\metre$ and $\sigma = \unit{7.5}\centi\metre$ in our illumination function. 
  }
  \label{fig:goal1-real}
  \Description[The geometry is recognizable through two corners]{There are five examples, in order: no geometry, a rectangle that is moved to the left, and to the right, a T-shaped geometry, and an upside-down T-shaped geometry. The image is brighter on the areas which have geometry.}
\end{figure*}

\paragraph{Looking around two corners}
In this experiment we use the scene shown in \fref{fig:goal1-real}a, where we image objects hidden behind two corners.
\new{The scene is made up of a diffuse surface} $\lM$ at a $\unit{45}{\degree}$ angle with respect to the relay surface, and several hidden geometries \NEW{$\lG$} (\fref{fig:goal1-real}b, top) oriented at $\unit{90}{\degree}$ with respect to the relay surface. Two occluders ensure that the \NEW{geometry} $\lG$ hidden around two corners is not directly visible from points $\xs \in \lS$, $\xl$, or the \new{capture hardware itself.}
For each geometry, we image \new{points $\xv$} in $\lV$ where the mirror images are produced by plane $\lM$, \new{using the confocal camera model \neww{(\eref{eq:RSD_freq_cc}) with the impulse response $H(\xl, \xs, t)$} to obtain $\fcc(\xv, t)$}.
\neww{\NEW{The frame at $t=0$ of} $\fcc(\xv, t)$} \NEW{captures the} objects hidden around two corners (\fref{fig:goal1-real}b, bottom) by leveraging five-bounce specular paths $\xl\rarr\lM\rarr\lG\rarr\lM\rarr\lS$ \new{in the computational domain}.
The resulting images appear blurrier than single-corner reconstructions since the mirror behavior at NLOS imaging frequencies is not perfectly specular (\sref{sec:discussion}).

\paragraph{\neww{Capture noise}}
\neww{Previous NLOS imaging methods that relied on single-pixel SPAD sensors suffered from low signal-to-noise ratio, requiring long capture times.
The implementation of gated SPAD array sensors \cite{riccardoFastGated16162022}, which we use in our work, significantly mitigates this issue, and can enable imaging speeds of up to five frames per second \cite{nam2021low}.
While the signal degrades with the number of bounces,
in our experiments we observed only minor changes (noise) in our computed images over multiple measurements of the same scene, for the computational wavelengths previously specified. This suggests that our imaging procedures are mainly affected by other limiting factors (e.g., surface size and reflectance) than by capture noise; in fact, in our work we had to lower the power of the laser to prevent overexposing our SPAD array sensor.
Additionally, we compared the photon count of our two-corner experiments (\fref{fig:goal1-real}) based on the fifth bounce, and their third-bounce counterparts with the target object $\lG$ placed at the mirror location $\lGp$ in a single-corner configuration. Under the same exposure time, the total photons captured in our fifth-bounce setups is \NEW{an order of magnitude higher}
than their third-bounce counterparts (\NEW{around $10^9$ and $10^8$ photons, respectively}), which suggests third- and fifth-bounce setups are similar in terms of capture noise when imaging similar regions of the hidden scene.}

\section{Discussion and future work}
\label{sec:discussion}

We have established a connection between the surface reflectance defined by well-known wave propagation principles and a wave-based NLOS imaging formulation, showing how diffuse planar surfaces become virtual mirrors at NLOS imaging wavelengths. We have then introduced \new{\newww{a procedure} to address the missing-cone problem}: by analyzing mirror images produced by other \NEW{virtual mirrors}, and then showing how to directly \NEW{image} such surfaces using secondary apertures. \new{Moreover, our insights \neww{have allowed} us} to image objects hidden behind two corners by imaging the space behind \NEW{virtual mirrors}, where we \neww{have observed} mirror images of objects hidden around two corners.

\paragraph{\neww{Mirror reflections under existing imaging methods}}
\neww{
In our work, we have showed how to image mirror reflections of different scene elements to address current limitations of NLOS imaging methods. 
To image objects around two corners, our key idea is to reason about the location of the imaged volume based on our analysis of mirror-like behavior of planar surfaces in the computational domain.
The evaluation of the confocal camera model at $t=0$ (which we use to obtain the image of the object hidden behind two corners) is equivalent to third-bounce imaging used by existing single-corner NLOS imaging methods.  
We have described our procedure using the wave-based phasor-field formulation of such imaging model.
Nevertheless, our methodology and observations could, in principle, generalize to existing single-corner methods that use similar models in order to extend them to two-corner scenes, providing an interesting path for future research.
}

\paragraph{\new{Fourth-bounce assumptions and higher-order bounces}}
\new{To address the missing-cone problem, we have used fourth-bounce illumination paths. Looking at \fref{fig:goal2-overview}a, \neww{the scene} has to meet two conditions to be able to compute an image of $\lG$, which is inside the null-reconstruction space of third-bounce methods.
First, the hidden scene must contain another surface that is not in the null-reconstruction space for third-bounce methods. In our scene, the surface $\lM$ has this purpose, which then can be used as a secondary aperture.
Second, there must exist a \NEW{four}-bounce path that reaches both the \neww{surface $\lM$ where the secondary aperture is located,} and the target surface $\lG$. This fourth bounce must be able to reach $\lS$ when following the specular bounce direction in the computational NLOS imaging domain, else both surfaces would be in the null-reconstruction space of fourth-bounce methods too.
Note that imaging surfaces with third-bounce illumination already requires similar assumptions. This could in principle generalize to fifth- or even higher-order bounces, allowing to create additional \neww{higher-order} apertures to observe further into hidden scenes.
A more thorough exploration of the potential of these \neww{higher-order} apertures is thus an interesting avenue of future work.}

\paragraph{\new{Fifth-bounce assumptions and multiple virtual mirrors.}}
\new{\neww{We have demonstrated how to image a surface $\lG$ hidden around two corners from fifth-bounce illumination, using a diffuse surface $\lM$ as a virtual mirror. For this to work, light from the illuminated point $\xl$ has to follow specular paths in the computational domain} that must reach $\lG$ after one bounce on $\lM$, and must reflect back to the aperture $\lS$ after another bounce on $\lM$, yielding a five-bounce path $\xl \rarr \lM \rarr \lG \rarr \lM \rarr \lS$. Thus, imaging $\lG$ depends on the location and orientation of $\lM$ with respect to $\lG$ and $\lS$. Note that this is no different from classic third-bounce NLOS setups, where objects must be located and oriented in regions outside of the null-reconstruction space to be imaged.
Our method could in principle generalize to more cluttered scenarios, where specular reflections between different planar surfaces would increase the coverage of NLOS imaging. To explore this, an exhaustive analysis of the connection between imaging wavelength, surface size and features, and their reflectance properties at different imaging frequencies would be necessary.}

\paragraph{\new{Coverage of the missing-cone problem using higher-order bounces}}
\new{In our work we have demonstrated how to image surfaces inside the null-reconstruction space of third-bounce methods. However, as can be seen in \fref{fig:goal2-results} (simulated) and \fref{fig:goal2-real} (captured), only a part of the surface $\lG$ reflects fourth-bounce illumination towards $\lS$ in the computational domain. Thus, some parts of $\lG$ remain inside the null-reconstruction space of our fourth-bounce imaging procedure. The third-bounce analysis of the missing-cone problem by \citet{Liu2019analysis} concludes that the visibility of a point in the scene only depends on the position of points on the visible relay surface. This is not the case for the null-reconstruction space of higher-order imaging methods, where the visibility of a point in the scene also depends on other hidden scene elements. Thus, a thorough analysis of the coverage of the missing-cone is an open challenging contribution in NLOS imaging regarded as future work.}

\paragraph{Mirror behavior} While our experiments showed that diffuse planar surfaces produce specular reflections at NLOS imaging frequencies, these reflections do not follow exactly a delta function due to diffraction effects. This happens likely because the \new{surfaces} we use are not much larger than the wavelength of the computational wave. Image quality therefore depends significantly on the size and position of the mirror surface. In our experimental results\new{, we observed that reflections through such diffuse surfaces appear blurry, mainly due to diffraction artifacts.}
\new{
\neww{Finally, some higher-order paths may have the same time of flight as third-bounce paths, and thus are coupled in the impulse response $H(\xl, \xs, t)$ introducing undesired artifacts in the imaging process, which is the case already for all existing NLOS imaging methods.
To what extent diffraction, imperfect mirror behavior, noise, and coupling between bounces enter this problem is an interesting topic for future research.}}

\new{In conclusion, our virtual mirrors framework addresses two of the most limiting problems of current NLOS imaging algorithms, leveraging fourth- and fifth-bounce illumination to compute images of surfaces inside the null-reconstruction space of existing methods, and even hidden behind two corners. We hope that our work spurs further research in this direction to explore the full potential of the field.}

\begin{acks}
We want to thank the anonymous reviewers for their time and insightful comments,
and the members of the Graphics and Imaging Lab for their help with the manuscript.
Our work was funded by the European Union's European Defense Fund Program through the ENLIGHTEN project under grant agreement No. 101103242, by the Gobierno de Aragón (Departamento de Ciencia, Universidad y Sociedad del Conocimiento) through project BLINDSIGHT (ref. LMP30\_21), by MCIN/AEI/10.13039/501100011033 through Project PID2019-105004GB-I00,
by the Air Force Office for Scientific Research (FA9550-21-1-0341), and by the National Science Foundation (1846884).
Additionally, Diego Royo was supported by a Gobierno de Aragón predoctoral grant.

\end{acks}

\bibliographystyle{ACM-Reference-Format}
\bibliography{bibliography.bib}


\end{document}